\documentclass[aip,jap,graphicx]{revtex4-1} 

\usepackage{amsmath} 
\usepackage{array} 
\newcolumntype{C}[1]{>{\centering\arraybackslash}p{#1}} 
\usepackage{makecell} 
\usepackage{amsmath, amssymb, graphicx, array, booktabs, caption}
\usepackage[margin=1in]{geometry}
\usepackage{makecell}
\usepackage{multirow}
\usepackage{hyperref}
\usepackage{siunitx}
\usepackage{graphicx}
\usepackage{booktabs} 
\usepackage{hyperref}
\usepackage[nameinlink]{cleveref}
\usepackage{verbatim}
\usepackage{subfig}
\usepackage{adjustbox}

\usepackage{color,soul}

\graphicspath{{./figures/}}

\Crefname{equation}{Eq.}{Eqs.}
\Crefname{figure}{FIG.}{Figs.}
\Crefname{tabular}{TAB.}{Tabs.}
\Crefname{table}{TAB.}{Tabs.}

\draft 

\begin{document}






\title[AI for Shear Bands]{A Physics-Aware Deep Learning Model for Shear Band Formation Around Collapsing Pores in Shocked Reactive Materials}
\author{Xinlun Cheng}
 \affiliation{School of Data Science, University of Virginia}
\author{Bingzhe Chen}
 \affiliation{School of Data Science, University of Virginia}
\author{Joseph Choi}
 \affiliation{School of Data Science, University of Virginia}
\author{Yen T. Nguyen}
 \affiliation{Department of Mechanical Engineering, University of Iowa}
\author{Pradeep Seshadri}
 \affiliation{Department of Mechanical Engineering, University of Iowa}
\author{Mayank Verma}
 \affiliation{Department of Mechanical Engineering, University of Iowa}
\author{H.S. Udaykumar}
 \affiliation{Department of Mechanical Engineering, University of Iowa}
\author{Stephen Baek}
 \email{mwn4yc@virginia.edu}
 \affiliation{School of Data Science, University of Virginia}
 \affiliation{Department of Mechanical and Aerospace Engineering, University of Virginia}

\begin{abstract}
Modeling shock-to-detonation phenomena in energetic materials (EM) requires capturing complex physical processes such as strong shocks, rapid changes in microstructural morphology, and nonlinear dynamics of chemical reactions fronts. These processes participate in energy localization at hotspots, which initiate chemical energy release leading to detonation. This study addresses the formation of hotspots in crystalline EMs subjected to weak-to-moderate shock loading which, despite its critical relevance to the safe storage and handling of EMs, remains underexplored compared to the well-studied strong shock conditions. To overcome the computational challenges associated with direct numerical simulations, we advance the Physics-Aware Recurrent Convolutional Neural Network (PARCv2), which has been shown to be capable of predicting strong shock responses in EMs. We improved the architecture of PARCv2 to rapidly predict shear localizations and plastic heating which play important roles in the weak-to-moderate shock regime. PARCv2 is benchmarked against two widely used physics-informed models, namely Fourier neural operator (FNO) and neural ODE; we demonstrate its superior performance in capturing the spatiotemporal dynamics of shear band formation. While all models exhibit certain failure modes, our findings underscore the importance of domain-specific considerations in developing robust AI-accelerated simulation tools for reactive materials.
\end{abstract}


\maketitle 

\textbf{Topics: } Deep learning, Artificial intelligence, Machine learning, Energetic materials, Shear modulus

\section{Introduction}
Shock initiation of energetic materials (EM) results from a combination of physico-chemical mechanisms that convert the mechanical energy of a shock into chemical energy release.\citep{field1992hot} When the timescale of this chemical energy release becomes comparable to the shock passage time, a detonation wave can form, leading to rapid and catastrophic energy release. Ensuring the safety of EMs therefore requires careful design and control of the energy release process during storage, transportation, or development of EMs. Typical EMs are composed of mixtures of organic energetic crystals (e.g., HMX), polymeric binders, and other additives.\citep{handley2018understanding} The microstructures of processed EMs are inherently heterogeneous, with energetic crystals and additives interspersed with unavoidable flaws, defects, and porosity. These defect sites act as zones of energy localization and initial chemical reactions when a shock propagates through the heterogeneous energetic material (HE).\citep{handley2018understanding} Such energy localization sites are known as hotspots\citep{field1992bhot}, and the ignition and growth of these hotspots play a crucial role in the shock-to-detonation transition (SDT) or the deflagration-to-detonation transition (DDT). A variety of mechanisms have been identified for hotspot formation.\citep{tarver1996critical,springer2017effects,springer2018modeling,rai2019void} Under strong shock conditions --- where shock pressures are in the range of several gigapascal (GPa), far exceeding the Hugoniot Elastic Limit of the HE --- the predominant hotspot mechanism is pore collapse.\citep{springer2017effects} However, at lower shock strengths, plastic dissipation begins to play a role, and shear bands appear to contribute to energy localization.\citep{coffey1987initiation,hamilton2024high} Under even weaker loading conditions, mechanisms such as friction between crystals can also play a role.

While the strong shock regime and resulting pore collapse have been widely studied, the intermediate shock strength range—where plasticity (deviatoric stress) competes with pore compression and collapse (pressure effects)—remains underexplored. This is largely due to the lack of simulation models that can accurately capture the dynamics of shear bands in a physically consistent manner. Only recently have works begun to address this gap, and the development of appropriate material models is still ongoing.\citep{pereverzev2020elastic,sewell2003molecular,nguyen2024continuum,herrin2024pore}

In the weak-to-moderate shock regime, pore collapse is coupled with plastic dissipation and localization in shear bands. The resulting hotspot shapes and temperature distributions are more complex and difficult to predict. The initiation and evolution of shear bands have been simulated using molecular dynamics (MD).\citep{sewell2003molecular,pereverzev2020elastic,chitsazi2020molecular} As shock strength increases, the density of shear bands rises, significantly affecting the plastic response of the EM.\citep{nguyen2024continuum} At the continuum level, modeling the shear- and shear-rate-dependent dynamics of shear bands and their impact on energy localization remains a challenge. In recent work, \citet{herrin2024pore} proposed an atomistically-consistent modified Johnson-Cook material model for HMX and RDX materials. They demonstrated that continuum simulations of hotspot formation over a wide range of shock strengths show good agreement with corresponding atomistic calculations. While continuum simulations greatly reduce computational cost (O(1) CPU hour on a computing cluster) compared to atomistic simulations (O($10^3$) CPU hours on computing clusters), performing ensembles of simulations required for constructing microstructure-informed reactive burn models remains computationally intensive.\citep{choi2023artificial}

In recent years, advances in artificial intelligence (AI) techniques have shown promise in significantly accelerating simulations, potentially predicting shock-generated hotspot formation within seconds on a GPU workstation. In general, physics-informed machine learning (PIML) models have demonstrated significantly higher accuracy than traditional physics-agnostic models, especially for complex dynamical systems in fluid flows and solid mechanics.\citep{karniadakis2021physics} An early and influential class of such methods is Physics-Informed Neural Networks (PINNs).\citep{raissi2019physics} In this approach, a deep neural network (DNN) approximates the solution of governing equations as a function of spatial coordinates ($x$, $y$, $z$) and time ($t$) variables. Automatic differentiation\citep{baydin2018automatic} is employed to compute spatial and temporal derivatives in the governing equations, and the residuals of these equations are minimized during PINN training. However, the nature of PINNs --- fitting a highly nonlinear function from space-time coordinates to equation solutions while minimizing a combination of equation residuals and data loss --- makes it difficult to generalize to new initial and/or boundary conditions, even when the governing equations remain unchanged. 

To address this limitation, alternative approaches such as neural operators (NOs) have been explored.\citep{kovachki2023neural} NOs learn mappings between function space operators and have demonstrated improved generalizability. Variants such as the Fourier Neural Operator (FNO)\citep{li2020fourier} and Physics-Informed Neural Operator (PINO)\citep{li2024physics} further enhance this generalizability to ``nearby" physical systems with different initial conditions. Another class of PIML models is Neural Ordinary Differential Equations (NODEs)\citep{chen2018neural}, where a neural network approximates the mapping between the system state at time $t$ and its time derivative. The predicted derivatives are then numerically integrated to obtain the state at the next time step, $t+\Delta t$. The network is not trained on ground-truth time derivatives; instead, the loss is computed based on the predicted states, and gradients are backpropagated through the numerical integrator --- either directly or via the adjoint method --- to update the network weights. This approach has seen widespread use beyond physical system dynamics, including cases where governing equations are unknown\citep{grathwohl2019ffjord,du2022flow}, or where inference across varying time steps is essential\citep{de2019gru}.

AI-accelerated or AI-supported simulations are also emerging in EM research, as summarized in the review articles by \citet{nguyen2023challenges}, \citet{choi2023artificial} and \citet{udaykumar2024impact}. Previous research was focused on generation of synthetic microstructures \citep{chun2020deep,nguyen2022synthesizing,roy2022meso,fang2025heterogeneous}, and predicting the meso- and/or micro-scale evolution of EM under a strong shock loading regime \citep{nguyen2023physics,nguyen2023parc}. For example, a novel deep learning architecture known as the Physics-Aware Recurrent Convolutional Neural Network (PARCv2; \citet{nguyen2024parcv2}), in which the structure of the advection-diffusion-reaction equations is embedded into the network design, has successfully modeled pore collapse under strong shock loading conditions. However, PARCv2's performance in predicting the dynamics of pore collapse under weak-to-moderate shock conditions was been found to be less satisfactory. Direct application of PARCv2 with its original architecture leads to deformation of pore geometry, spurious high-temperature regions near boundaries, and numerical instability within 20 roll-out steps across a broad range of impact velocities. The application of machine learning to directly model EM under weak-to-modest shock conditions has been largely understudied, and this work represents a pioneering work in identifying, investigating and proposing solutions to the challenges associated with this problem, namely spectral bias, boundary condition enforcement, and performance deterioration under low impact velocity.

In this paper, we present substantial improvements to the PARCv2 architecture for modeling shear band formation in materials subjected to weak-to-moderate shock loading. In \Cref{sec:data}, we describe data generation, model setup, and training strategies. In \Cref{sec:results}, we present our findings, demonstrating that while PARCv2 achieves state-of-the-art accuracy compared to two other popular PIML models, all three deep learning models struggle with extrapolation and capturing finer details. In \Cref{sec:discussion}, we analyze the likely causes of these limitations and propose several refinements to the training strategy and network architecture. We investigated whether spectral bias reduction methods proven successful in computer vision community can translate to similar improvements on physics modeling problem. We examined the theoretical and computational foundations for enforcing various types of boundary conditions through different padding operations in convolutional neural networks. We also broke down the performance decrease in low velocity cases into three likely reasons and unique challenges associated with modeling complex dynamical processes. Finally, in \Cref{sec:conclusion}, we draw our conclusions and outline directions for future work.

\section{Data Generation}\label{sec:data}
\subsection{Simulation Setup and Dataset}
\begin{figure}
    \centering
    \includegraphics[width=0.5\columnwidth]{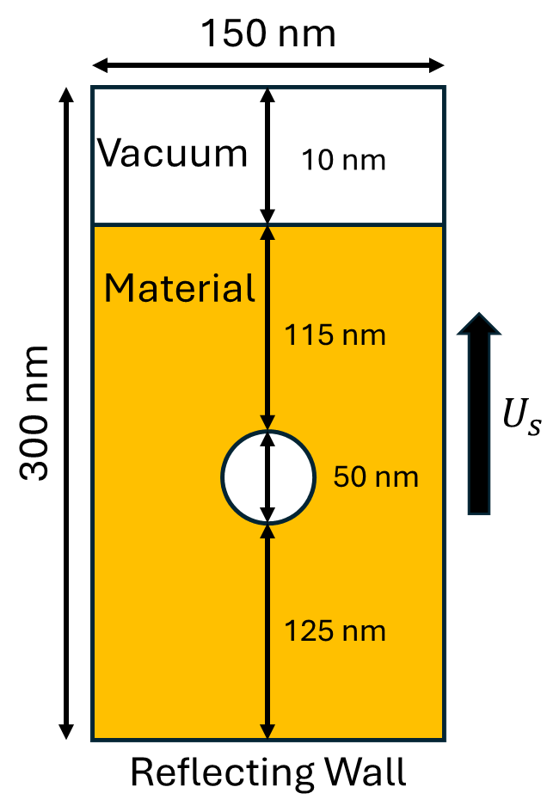}
    \caption{Simulation configuration for this work. A rectangular block of RDX with a single $50\,\mathrm{nm}$ pore impacts a rigid lower wall, generating a supported shock wave that propagates upward and initiates pore collapse and shear localization.}\label{fig:domain_sketch}
\end{figure}

Simulation data are acquired from \citet{herrin2024pore}, capturing the dynamics of pore collapse under the influence of shock waves. The calculations are performed using the continuum dynamics code SCIMITAR3D\citep{rai2018three}, in the configuration shown in \Cref{fig:domain_sketch}, in which a rectangular block of RDX contanining an initial circular shaped pore of diameter $D_{void}=50 \, nm$ is subjected to normal-incidence shock loading in a reverse-ballistic configuration wherein the sample impacts with velocity $U_p$ onto a rigid and stationary wall at $y=0$. This results in a supported shock wave that initially travels with velocity $U_w$ in the frame of the stationary piston, anti-parallel to $U_p$. The shock velocity $U_s$ in the material frame (i.e., a reference frame moving with the sample) is $U_s = U_w + |U_p|$. For governing equations and simulation setup details, readers are referred to the Appendix of this paper, as well as the supplementary materials of \citet{nguyen2024continuum} and \citet{herrin2024pore}, which provide extended formulations and MD-informed model parameters.

The simulation data are rasterized onto a $128 \times 256$ grid, corresponding to a spatial resolution of $\Delta x = 1.1719$ nanometers per pixel (nm/pixel), with a temporal resolution of $\Delta t = 2.5$ picoseconds (ps) between adjacent snapshots, starting from the initial condition at $t = 0$. The training-validation sets cover initial impact velocities from 720 m/s to 2880 m/s in 20 m/s increments. The training and validation sets are then randomly split from the combined set, yielding 70 training simulations and 18 validation simulations.

The hold-out testing set includes all cases with initial impact velocities divisible by 100 within the aforementioned range, as well as cases in the ranges of 500–700 m/s and 2900–3080 m/s. There are 26 simulations in this hold-out testing set. These cases are selected to assess both in-distribution interpolation and out-of-distribution extrapolation performance of the deep learning models. To ensure fair model comparison, the same training-validation-test split is used across all three deep learning models evaluated in this study.

\begin{figure}
    \centering
    \includegraphics[width=\linewidth]{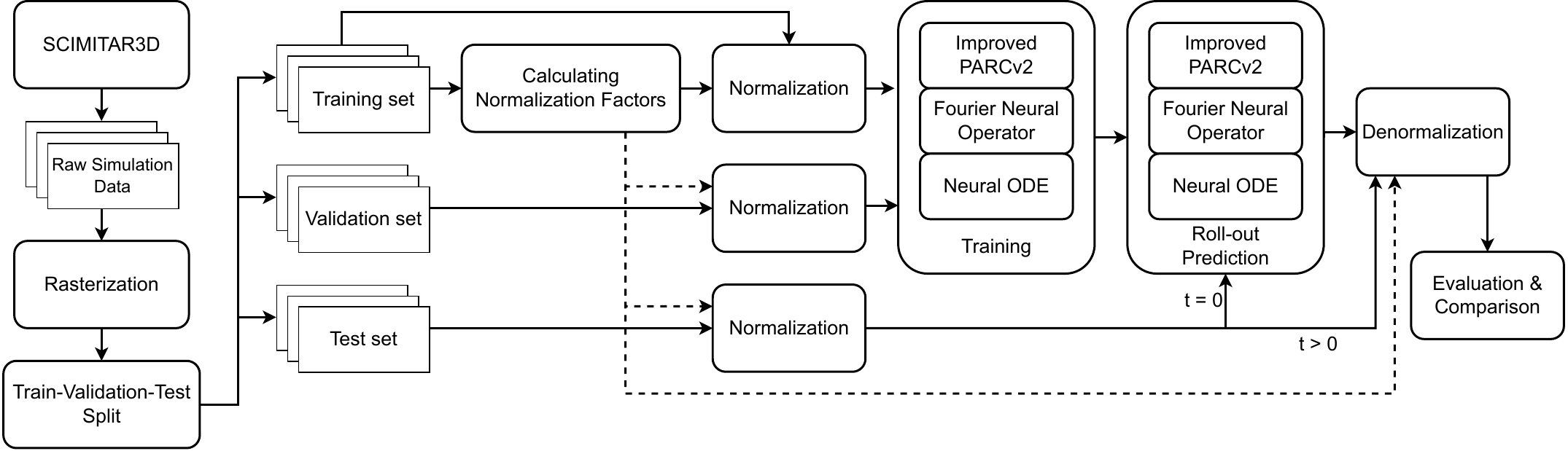}
    \caption{Flowchart of this work.}
    \label{fig:flow_chart}
\end{figure}

The dataset is normalized prior to training. Temperature and pressure channels are min–max normalized, while velocity fields are normalized as vectors such that each component lies between $-1$ and $1$. Pore morphology does not require normalization, as its values are inherently bounded between 0 (vacuum/pore) and 1 (material). Normalization constants are computed using only the training set. The input channels consist of temperature, pressure, pore morphology, and velocity fields, with outputs provided in the same configuration but at subsequent time steps. Because the dataset is uniformly spaced in time and none of the models explicitly depend on time, the actual timestep value does not affect prediction accuracy; we therefore supply a fixed $\Delta t = 0.1$ to all models. All three ML models are trained and evaluated with a single time step as input but, owing to their auto-regressive nature, can generate arbitrarily many future time steps as specified by the user. Model architecture, training hyperparameters and strategies for the three models are presented in the following sections, and their key differences are summarized in \Cref{tab:model_comparison}. Flowchart representing the entire workflow of this paper is presented in \Cref{fig:flow_chart}.

\begin{table}[!ht]
    \centering
    \begin{tabular}{c@{\hskip 0.2in}c@{\hskip 0.2in}c}
    \hline
    Model & Architecture Highlights & Physics Law Enforcement \\
    \hline
    Improved PARCv2 & Differentiator-integrator & Embedding governing eqn structures \\
    Fourier Neural Operator & Operator learning & Fourier mode operators \\
    Neural ODE & Differentiator-integrator & Learned from data \\
    \hline
    \end{tabular}
    \caption{A comparison of the ML models presented in this work.}
    \label{tab:model_comparison}
\end{table}

\subsection{Improved PARCv2}
We adopted a novel architecture, PARCv2, to address this problem due to its demonstrated capability to model complex transient physical phenomena, such as strong shock loading of complex material systems with non-linear reactions\citep{nguyen2024parcv2} and supersonic flow\citep{cheng2024physics}. The overall network design is similar to the architectures presented in the two cited works, but includes several notable improvements, which are detailed below:
\begin{itemize}
    \item In the original work, advection was computed using right derivative regardless of the velocity direction. We have updated this to use an upwind numerical scheme for physical consistency with the advection operator and improved stability.
    \item Boundary conditions in the original implementation were limited to zero padding. We have extended this to support constant, periodic, and zero-gradient boundary conditions. For this particular problem, we enforced zero-gradient boundary conditions across all boundaries of the domain.
    \item We adopt a multi-step curriculum training strategy, beginning with 3000 epochs of training to predict the next timestep using a constant learning rate of $10^{-4}$, followed by 1000 epochs of training to predict the next three timesteps using a learning rate of $10^{-5}$.
\end{itemize}
The advection of the five state variables --- temperature ($T$), pressure ($p$), microstructure ($\mu$), horizontal ($U$) and vertical ($V$) velocities --- is explicitly calculated using an upwind numerical scheme, while additional couplings between them are modeled using a U-Net\citep{ronneberger2015u}. The fourth-order Runge–Kutta (RK4\citep{runge1895numerische,kutta1901beitrag}) method is employed for temporal integration, and no data-driven integrator is used. The model contains approximately 15.7 million trainable parameters.

A mean absolute error (MAE, L1) loss function and the Adam optimizer\citep{kingma2017adam} are employed for both training stages, and no regularization techniques are applied. The model is trained on four Nvidia A100 GPUs using the PyTorch\citep{ansel2024pytorch} Distributed Data Parallel (DDP) framework, with a total wall-clock training time of 72 hours. The improved PARCv2 package and project code are publicly available at \url{https://github.com/baeklab/PARCtorch} and \url{https://github.com/chengxinlun/shear_bands}, respectively.

\subsection{Fourier Neural Operator}
Fourier Neural Operator (FNO) is a deep learning architecture designed to efficiently solve partial differential equations (PDEs) by learning mappings between function spaces. FNO is publicly available under the package \textit{neuraloperator}, which can be found at \url{https://github.com/neuraloperator/neuraloperator}; version 0.3.0 was used in this work. The architecture leverages Fourier transforms to operate in the frequency domain, enabling it to capture global structures more effectively than many other machine learning methods. FNO integrates neural networks with Fourier transforms to directly learn operators in the Fourier space, making it particularly effective for a range of computational fluid dynamics problems\citep{qi2024gabor,tran2021factorized,zhang2022fourier}, as spectral and pseudo-spectral method have seen wide usage of application in solving Poisson, Burgers, and incompressible Navier-Stokes equations\citep{qin2024toward}.

In this work, we employ Tensorized Fourier Neural Operator (TFNO2d) recursively in time to model the shear band formation process, following the recommendation of \citet{li2020fourier}, who advocated the use of recursive FNO2d over FNO3d in data-constrained scenarios. We use four Fourier blocks, retaining the first 24 Fourier modes due to computational resource limitations. Additionally, 78 channels are used for lifting, projection, and projection layers, resulting in a model with a number of trainable parameters comparable to the previously discussed PARCv2 model. All other architectural hyperparameters are kept at their default values.

The model is trained to perform single-step prediction for 10,000 epochs with a constant learning rate of $10^{-4}$. Attempts to extend training to predict the next three timesteps showed no significant improvement after 100 epochs and were therefore not pursued further.

\subsection{Neural ODE}
Contrary to PINN or FNO, neural ODE approximates the PDE system directly with a neural network and uses numerical integration methods to evolve from the predicted temporal derivatives of physical variables to their values at subsequent snapshots\citep{chen2018neural}. The trainable weights in the network can be updated either by backpropagating directly through the numerical integrator or via reverse-time integration of the adjoint equation, as the temporal derivatives are often difficult or impossible to measure directly.

Neural ODEs have gained popularity in the computer vision community for their ability to continuously transform one probability distribution into another (continuous normalizing flow\citep{grathwohl2019ffjord}) and are attractive for modeling physical systems due to their similarity to conventional numerical simulation routines.

In this work, we use a ResNet\citep{he2016deep} architecture with approximately 15.5 million trainable parameters to approximate the five temporal derivatives from a given input state. The predicted derivatives are then passed through an RK4 numerical integrator to obtain the predicted field values at the next snapshot. The input is first processed through two convolution layers to lift it to 460 channels, followed by four ResNet blocks. Each ResNet block contains two paths: one where the input undergoes a sequence of convolution–batch normalization–ReLU–convolution–batch normalization operations, and another where the input remains unchanged. The outputs of these two paths are added together and passed through a final ReLU activation to produce the output of the ResNet block. An additional $1 \times 1$ convolution is applied after the final ResNet block. All convolution operations use $3 \times 3$ kernels unless otherwise specified. This model has a similar number of trainable parameters to the PARCv2 model described earlier.

The model is trained on single-step prediction for 3000 epochs with a constant learning rate of $10^{-4}$. Attempts to train the model to predict the next three timesteps using a reduced learning rate did not yield significant improvements beyond 100 epochs, likely due to the more complex loss landscape and diminishing gradients encountered during backpropagation through the numerical integrator.

\section{Results}\label{sec:results}
In this section, we present the results from applying the three deep learning techniques described above to learn and predict the dynamics of pore collapse in the reverse ballistic impact configuration shown in \Cref{fig:domain_sketch}. A detailed study of the physics underlying this phenomenon has recently been presented in \citet{herrin2024pore}. The top row of \Cref{fig:temp_1800} illustrates a typical example of pore collapse for an impact velocity of 1800 m/s. Upon impact, a shock wave travels upward toward the pore, causing it to collapse. The primary features of the resulting hotspot include a high-temperature core and surrounding shear bands that exhibit energy localization and moderately elevated temperatures. Additionally, a blast wave propagates outward from the site of pore collapse, as seen in the pressure plots in the fourth column of the top row of \Cref{fig:pres_1800}.

The performance of each AI model is evaluated based on its ability to accurately reproduce the evolution of key field variables -- temperature, pressure, and velocity -- as well as the pore shape during collapse, relative to the ground truth. All results presented below are based on roll-out prediction over the test set, in which the models are provided only the initial conditions at $t = 0$ and tasked with predicting all subsequent snapshots for test cases that were not seen during training. The number of predicted steps varies depending on the initial impact velocity, ranging from 15 to 50 steps.

\subsection{Roll-out RMSE Comparison}\label{sec:rmse}
\begin{figure}
    \includegraphics[width=0.45\textwidth]{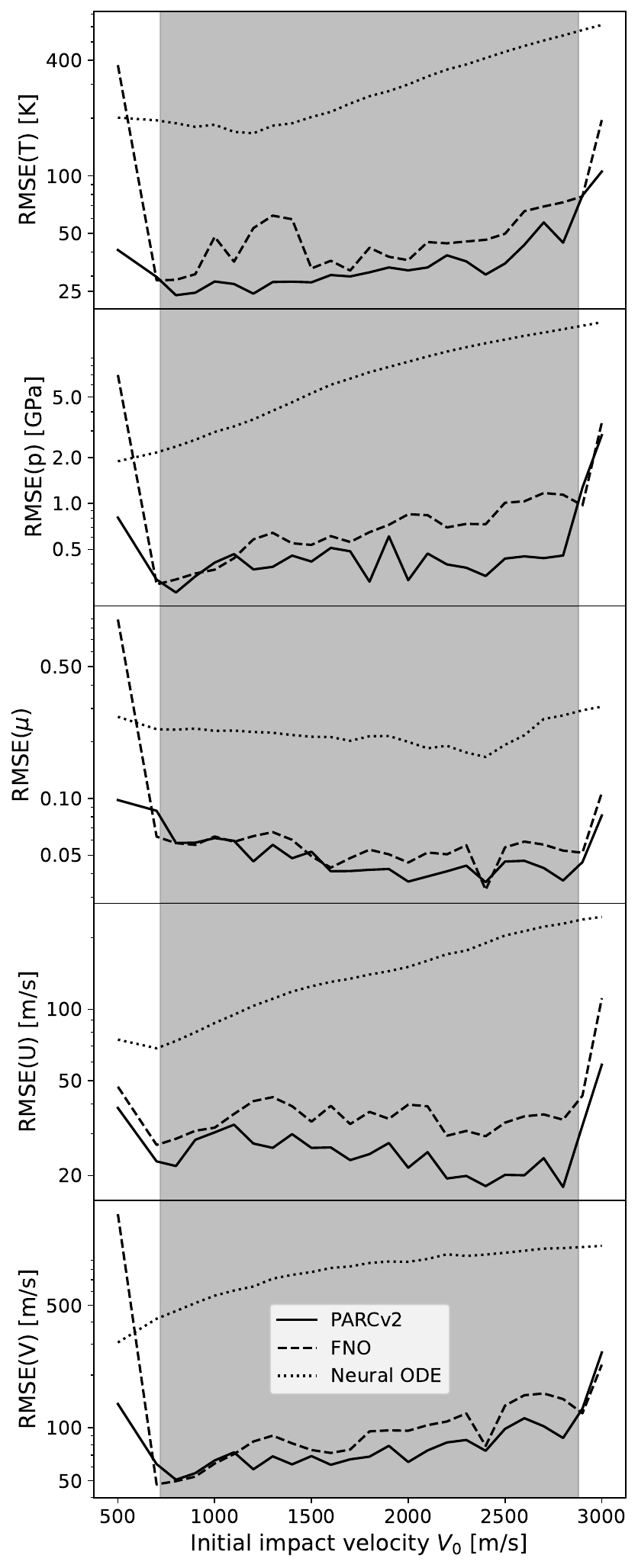}
    \caption{Initial impact velocity versus root mean squared error (RMSE) of model predictions Shaded areas indicates coverage of training set.}\label{fig:mean_mae}
\end{figure}

We first examine the model accuracy and compare the three models using root mean squared error (RMSE) against the ground truth (DNS simulation data in the test set). RMSE values reported are averaged over the entirely of the simulation for each channel for each impact velocity. As previous research indicates that PIML model accuracy is related to initial conditions (e.g., \citet{cheng2024physics}), we present the RMSE in \Cref{fig:mean_mae} as a function of initial impact velocities to examine whether a similar dependency exists in the modeling of shear bands. The results show that PARCv2 consistently achieves comparable or lower RMSE values regardless of the physical quantity predicted, indicating that it outperforms the other two models in terms of accuracy across all five fields. The reduction in RMSE achieved by PARCv2 is particularly notable in low-velocity extrapolation cases.

Interestingly, we do not observe significant improvement when comparing PARCv2 predictions to those of FNO for the pore profile or the horizontal (X-) component of the velocity field. We suspect that the pore profile evolution may not be sufficiently complex to fully demonstrate the advantages of PARCv2, while the patterns in the X-velocity field are highly disordered, leading to reduced accuracy in both models. Compared to the neural ODE, PARCv2 achieves nearly a tenfold decrease in RMSE across the five fields, indicating that the inclusion of explicitly calculated advection and diffusion information enables the model to better learn the physical system dynamics by directing attention to regions with strong advection and gradients.

Nevertheless, we find that even for PARCv2, the RMSE of temperature (25–50 K when within the training set coverage) is comparable to or sometimes greater than the temperature difference between shear bands and the surrounding material, suggesting that non-dominant shear bands may be absent from the predictions of all AI models. Given the physical importance of accurately predicting the evolution and formation of shear bands in this study, we now focus on a more detailed examination of the rollout-predicted fields in this context.

\subsection{Visual Examination of Roll-out Prediction}
\subsubsection{Interpolation}\label{sec:inter}
\begin{figure*}
    \includegraphics[width=\textwidth]{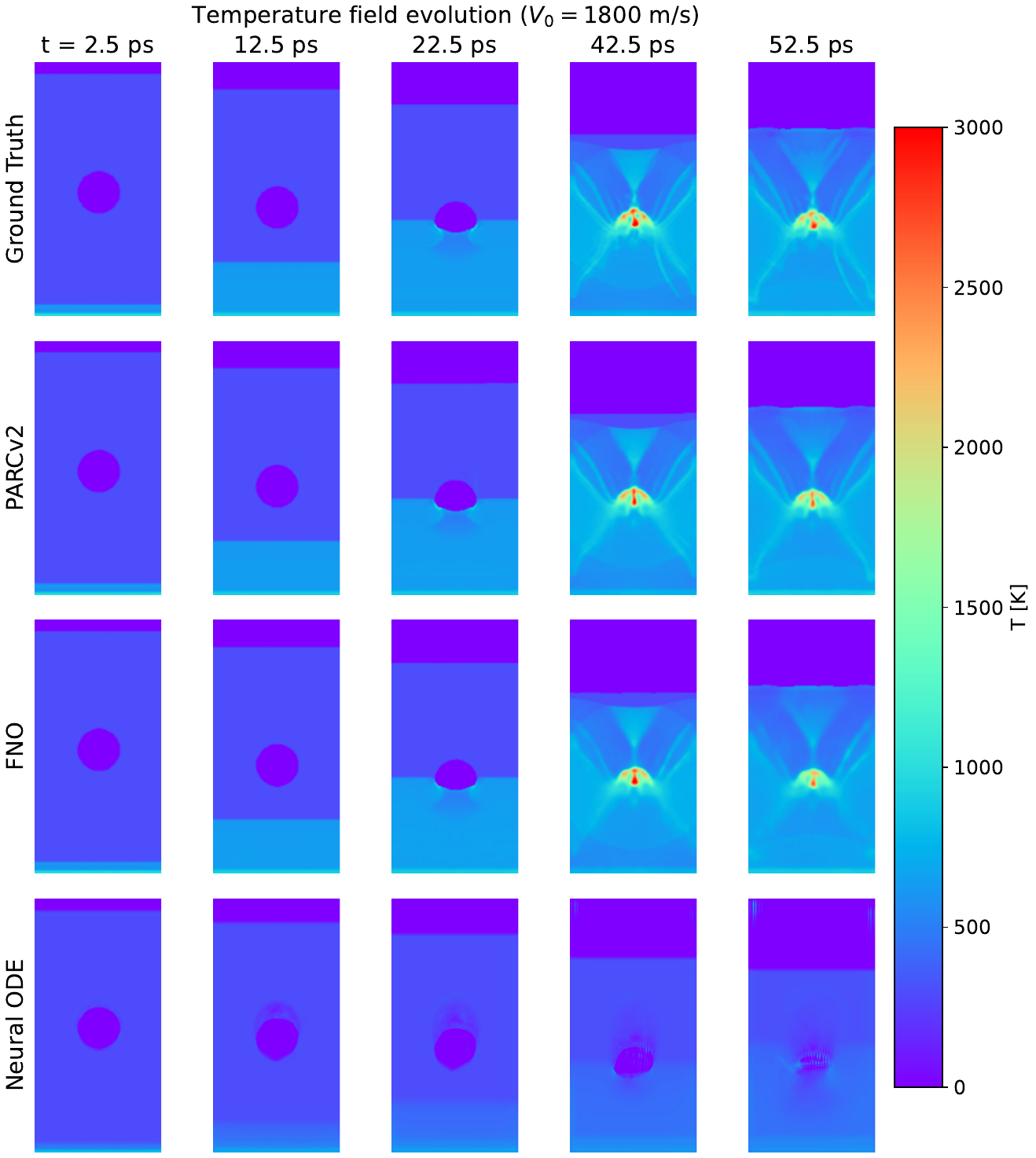}
    \caption{Temperature field evolution of initial velocity of 1800 m/s.}\label{fig:temp_1800}
\end{figure*}

\begin{figure*}
    \includegraphics[width=\textwidth]{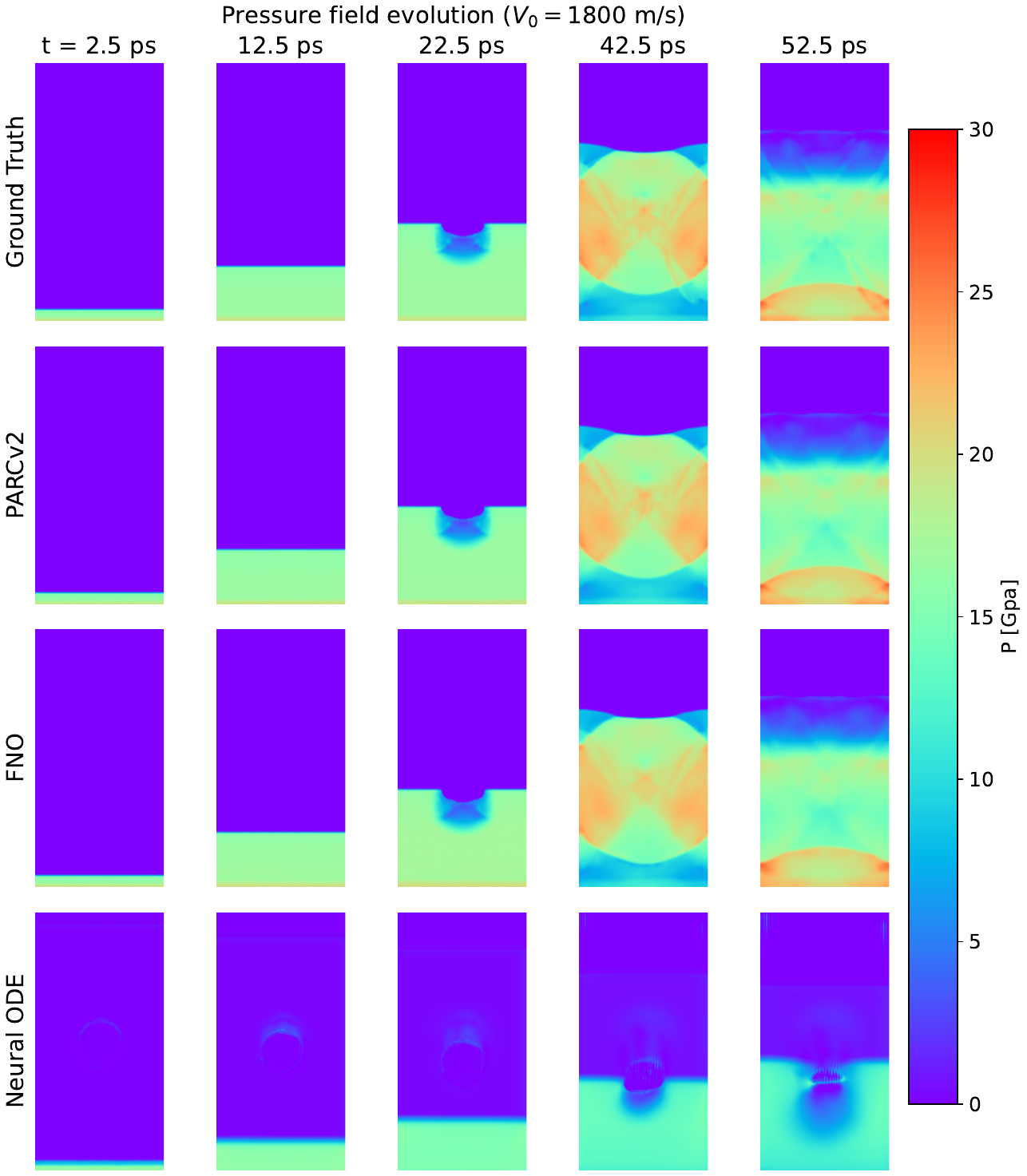}
    \caption{Pressure field evolution of initial velocity of 1800 m/s. }\label{fig:pres_1800}
\end{figure*}

\begin{figure*}
    \includegraphics[width=\textwidth]{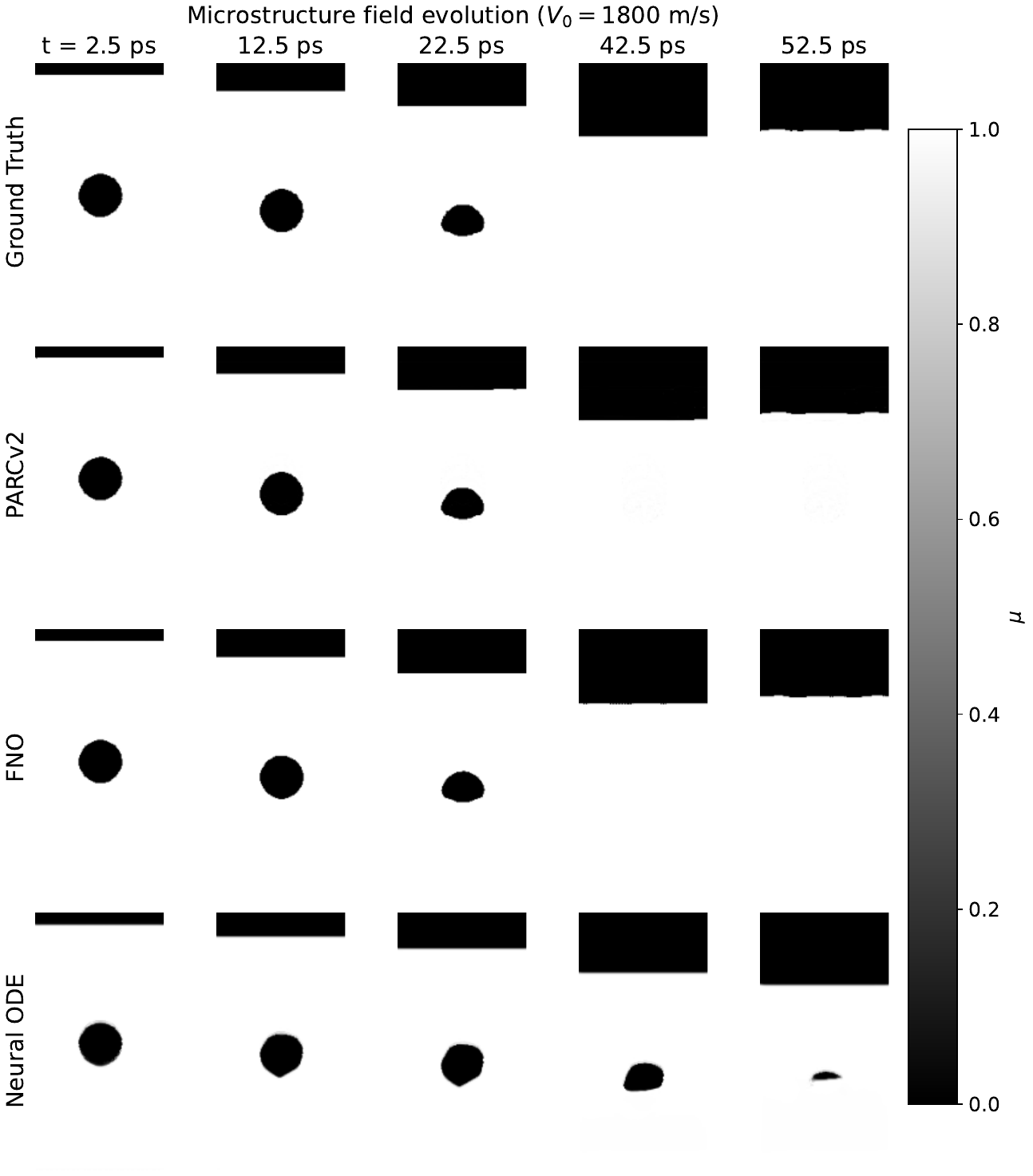}
    \caption{Microstructure field evolution of initial velocity of 1800 m/s. }\label{fig:micr_1800}
\end{figure*}

\begin{figure*}
    \includegraphics[width=\textwidth]{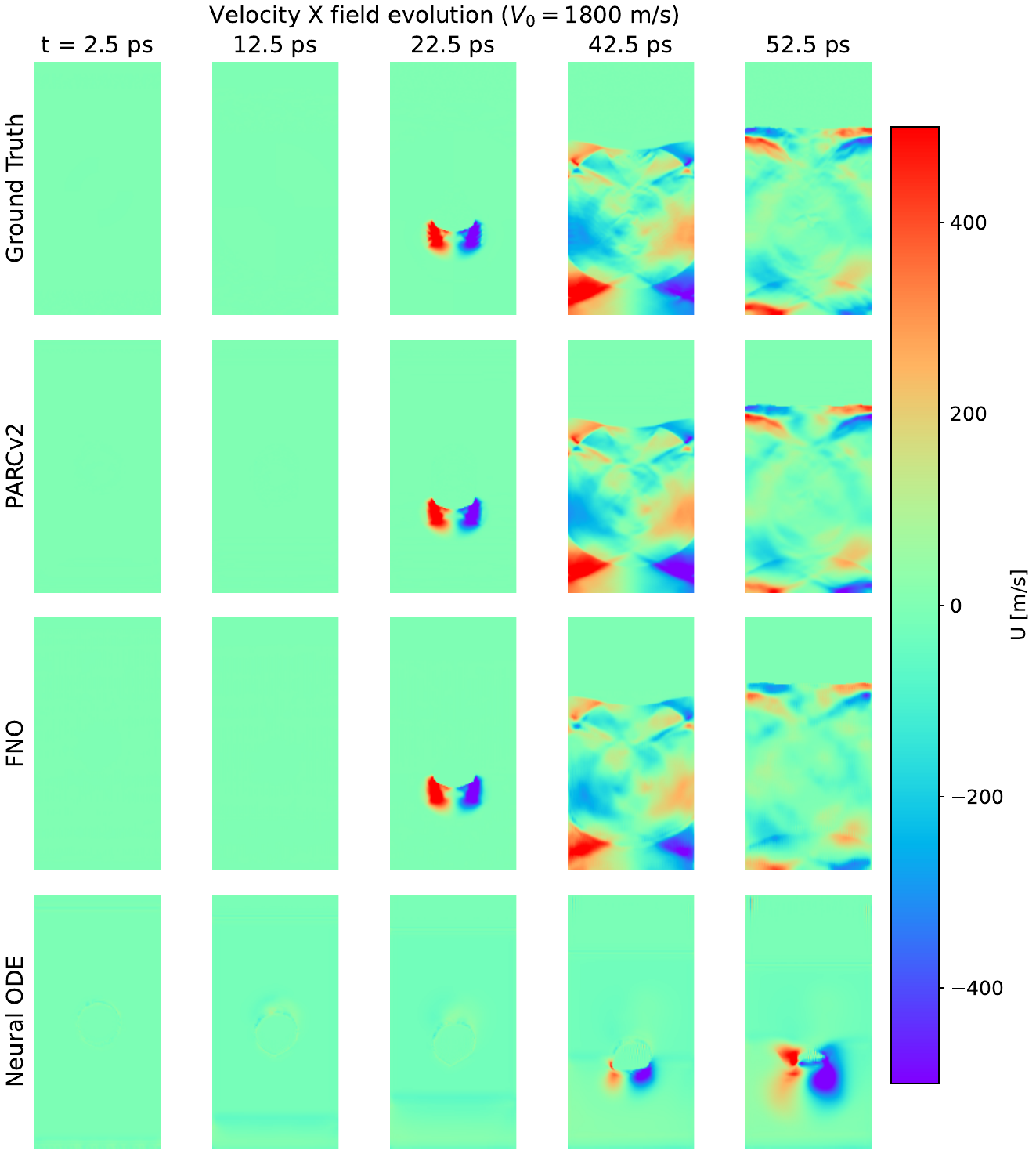}
    \caption{X-direction velocity field evolution of initial velocity of 1800 m/s. }\label{fig:xvel_1800}
\end{figure*}

\begin{figure*}
    \includegraphics[width=\textwidth]{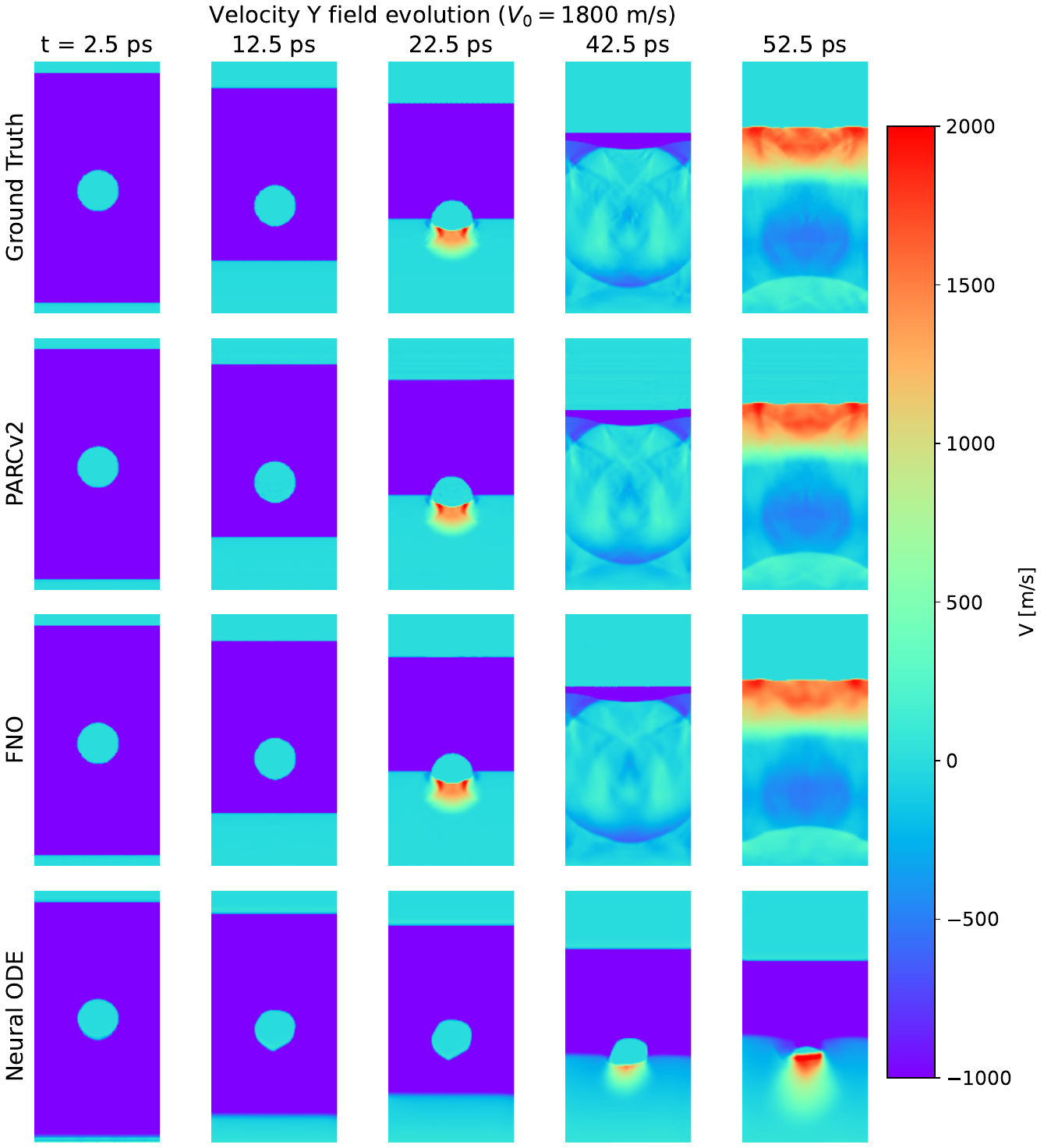}
    \caption{Y-direction velocity field evolution of initial velocity of 1800 m/s.}\label{fig:yvel_1800}
\end{figure*}

We begin our evaluation of model performance with test cases that fall within the training set coverage. \Cref{fig:temp_1800}---\Cref{fig:yvel_1800} show the ground truth and predicted evolution for an initial velocity of 1,800 m/s. Throughout the simulation, the predicted sequences from PARCv2 (second row) and FNO (third row) are in close agreement with the ground truth. Neural ODE with a ResNet differentiator (last row), on the other hand, fails to produce reasonable predictions. In the early stage of the simulation (first two columns), when the shock travels through the material before contacting the circular pore, there are no visible differences. During this stage, all three models correctly predict the trivial all-zero solution for the X-velocity, as shown in the first two columns of \Cref{fig:xvel_1800}.

During pore collapse (the third and fourth columns of \Cref{fig:temp_1800} --- \Cref{fig:yvel_1800}), while the overall patterns are well captured by both PARCv2 and FNO, some finer shear bands present in the ground truth pressure field --- such as the alternating blue-cyan regions just below the pore in the top panel of the third column of \Cref{fig:pres_1800} --- are smoothed out in the predictions, particularly in the bottom panel of the same column. A similar phenomenon is observed in the predicted X and Y velocity fields in the third and fourth columns of \Cref{fig:xvel_1800}. Nevertheless, the dominant shear bands predicted by PARCv2 are noticeably sharper than those produced by FNO, and therefore more closely match the ground truth.

After the pore has fully collapsed and a region of high temperature has developed (fourth column of \Cref{fig:temp_1800}), both PARCv2 and FNO successfully capture the shape of the high-temperature region (the central bird-shaped pattern), the shock heating phenomenon (a light blue circle centered on the pattern), and several of the most prominent shear bands (cyan bands radiating from the tips of the central structure). In the pressure field after pore collapse (\Cref{fig:pres_1800}), both models capture most of the complex shear bands (red bands radiating from the center) and the shock wave generated by the high-temperature region (green circle). However, the predicted pressure fields are less sharp than the ground truth. The earlier observation that PARCv2 produces sharper predictions that more closely resemble the ground truth remains valid at this stage.

Lastly, in the later stages of the simulation (last column of \Cref{fig:temp_1800}), when reverse ballistics dominate, the reflected pressure wave from the bottom boundary is predicted with high accuracy by both PARCv2 and FNO. At this point in the simulation, the X-velocity field becomes highly disordered, and the predicted field loses some finer details, as seen in \Cref{fig:yvel_1800}. However, the anti-symmetric pattern around the central vertical axis remains well preserved in the predictions from both PARCv2 and FNO. For the Y-velocity, we observe a behavior similar to that of the pressure field: the overall pattern closely resembles the ground truth, with prominent features --- such as shock waves generated by the central high-temperature region and those reflected from the bottom boundary --- being accurately predicted, though finer structures appear smeared out.


\begin{figure*}
    \includegraphics[width=\textwidth]{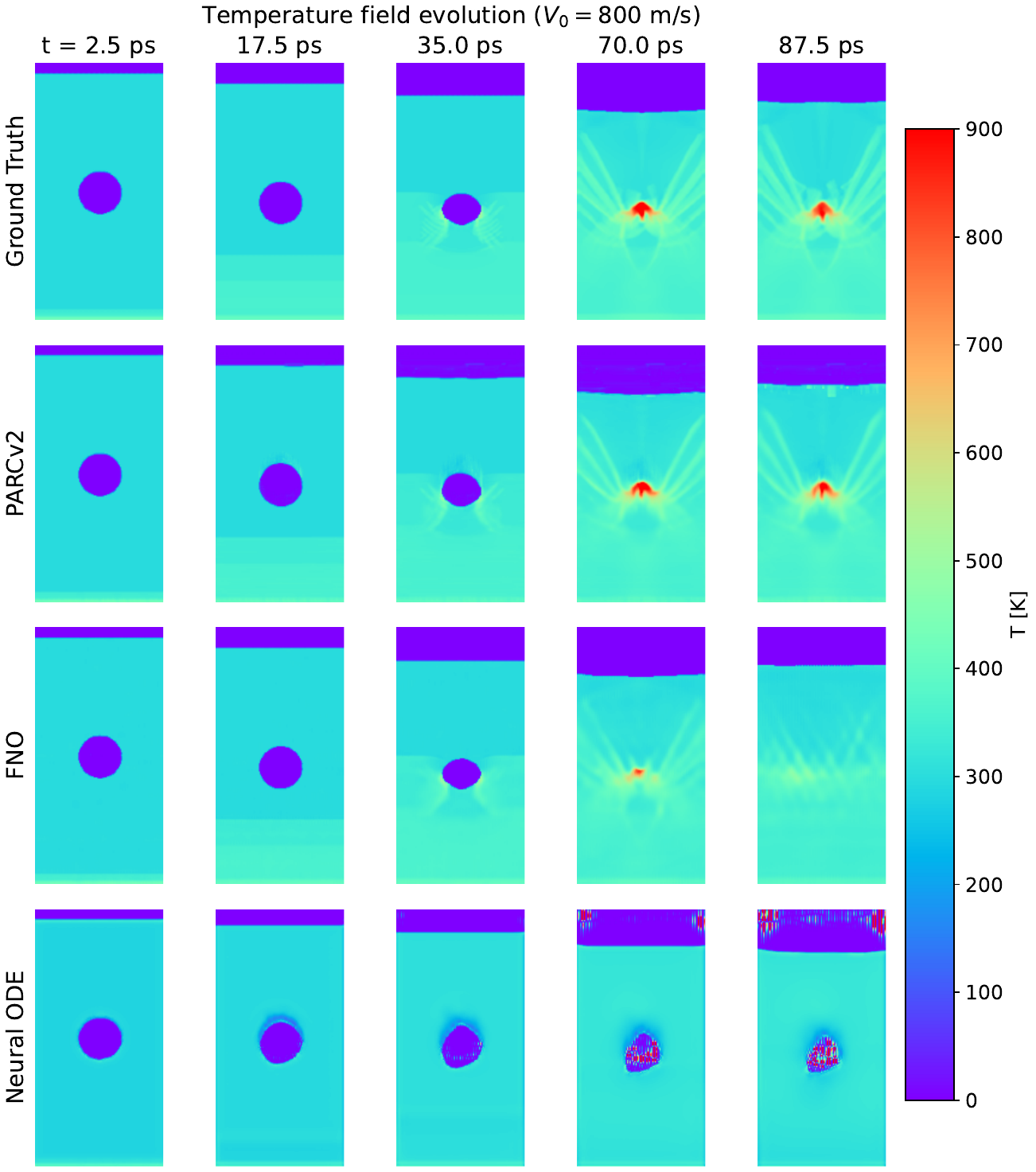}
    \caption{Temperature field evolution of initial velocity of 800 m/s.}\label{fig:temp_800}
\end{figure*}

\begin{figure*}
    \includegraphics[width=\textwidth]{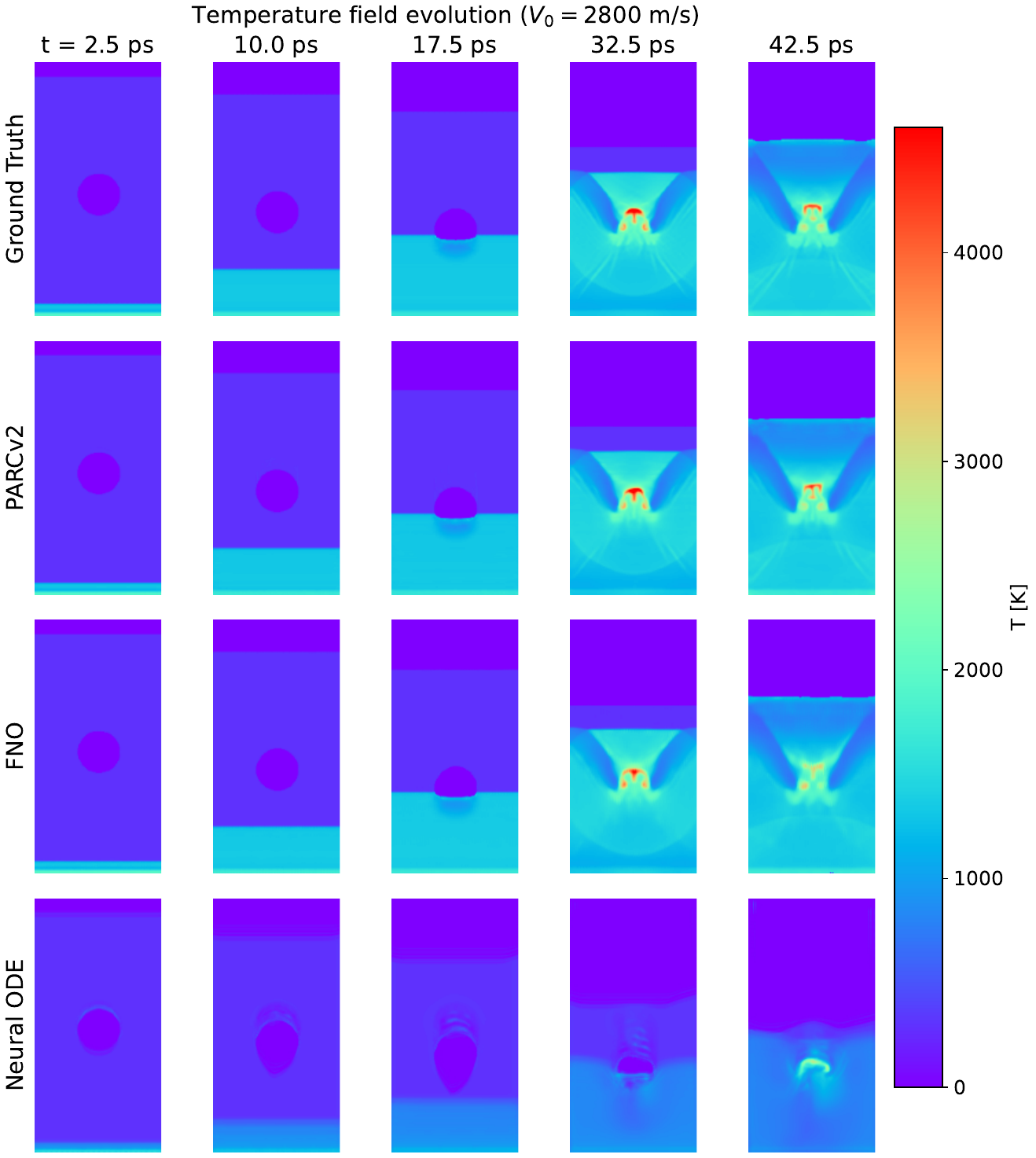}
    \caption{Temperature field evolution of initial velocity of 2800 m/s.}\label{fig:temp_2800}
\end{figure*}

We then visually examine the predictions at $V_0$=800 m/s and $V_0$=2,800 m/s in \Cref{fig:temp_800} and \Cref{fig:temp_2800}, respectively --- the lowest and highest velocity test cases within the training set coverage. For brevity, we present only the temperature fields, as the shear banding phenomenon is most prominent in this field.

For the 800 m/s impact velocity case, the PARCv2 prediction is satisfactory in that three dominant shear bands are accurately predicted at $t=70$ ps. However, the FNO prediction begins to deviate rapidly from the ground truth as the localized high-temperature region starts to form. This clearly reveals a limitation of commonly used PIML models: fast transient features are considerably more difficult to learn, likely due to their infrequent appearance in the training data. Indeed, the ResNet-based neural ODE is seen to fail in this case. Additionally, most of the weaker shear bands are completely smeared out in the predictions produced by all three methods.

For the 2,800 m/s impact velocity case, as shown in \Cref{fig:temp_2800}, PARCv2 outperforms FNO, particularly in the prediction of both the localized region of high temperature and the dominant shear bands. Although the quantitative RMSE loss shown in the top panel of \Cref{fig:mean_mae} indicates that the RMSE value for the 2800 m/s case is nearly twice that of the 800 m/s case, the predictions appear visually closer to the ground truth. We suspect that this is because the highest temperature values in the 800 m/s case are significantly lower in value than those in the 2800 m/s case, making deviations more noticeable in the lower-velocity predictions despite the lower RMSE. This finding not only underscores the importance of visually examining model predictions but also contrasts with typical numerical calculations, where high-velocity cases usually pose greater challenges for coarse-grid simulations. We provide further discussion on this behavior in the following sections.


\subsubsection{High Velocity Extrapolation}
\begin{figure*}
    \includegraphics[width=\textwidth]{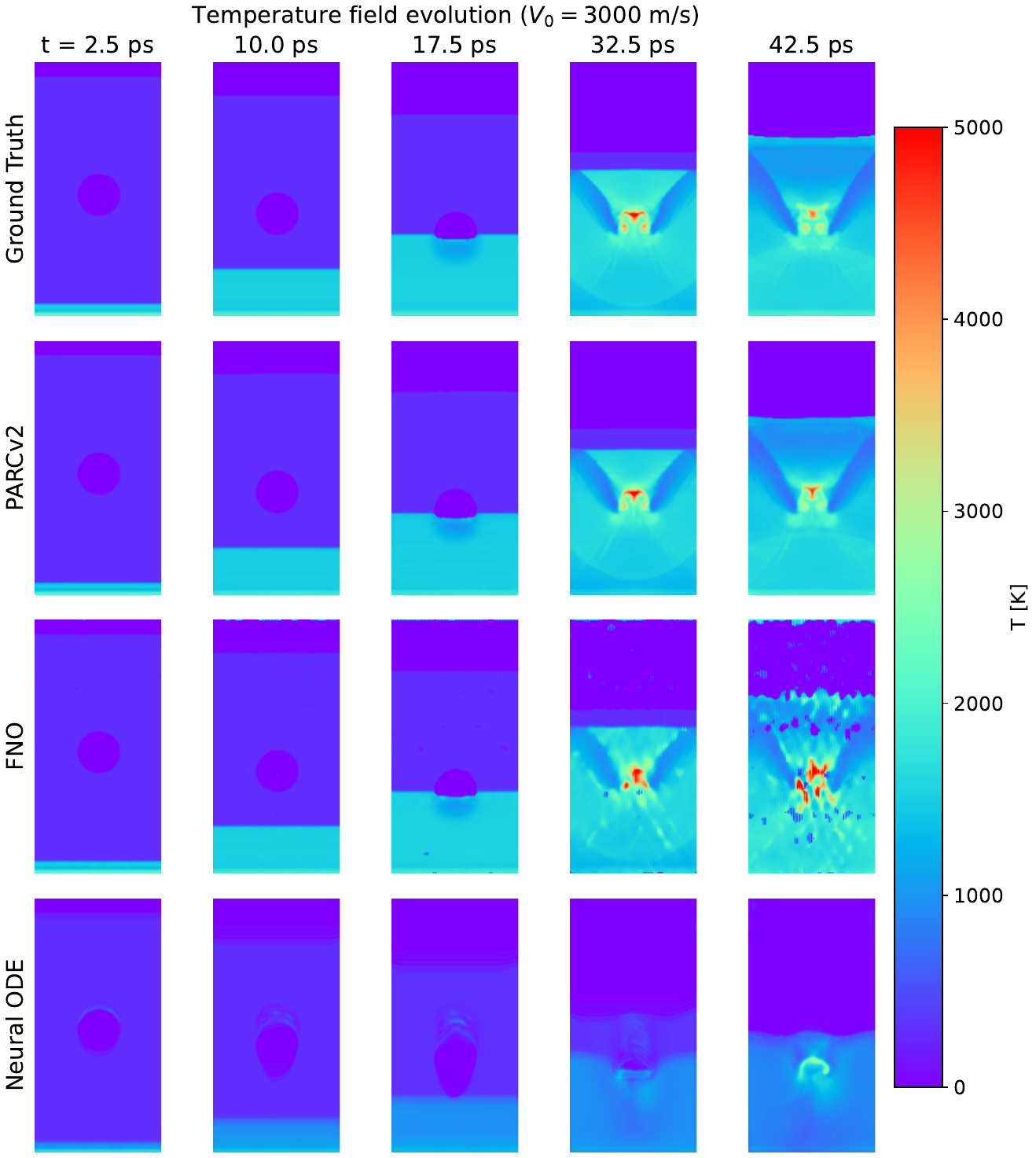}
    \caption{Temperature field evolution of initial velocity of 3000 m/s.}\label{fig:temp_3000}
\end{figure*}

A few test cases were specifically selected to evaluate the ability of the three models to extrapolate beyond the coverage of the training set. We begin with an examination of high-velocity extrapolation cases ($V_0 >$ 2,900 m/s). Quantitatively, we observed a sharp increase in RMSE values, as shown in \Cref{fig:mean_mae}. Examination of the predicted fields revealed no noticeable artifacts for PARCv2; the increased error is primarily due to inaccuracies in predicting the shape of the central high-temperature region and an overall blurriness in the predicted fields. In the later stages of the simulation (4th and 5th columns of \Cref{fig:temp_3000}), while the ground truth displays a blunt-shaped high-temperature region, the PARCv2 prediction retains a relatively sharp shape. Although the most prominent shear bands—wide blue bands radiating from the sides of the central high-temperature region—are well predicted, many smaller and weaker structures are lost in the prediction.

In contrast, FNO exhibits significant numerical artifacts beyond $t=32.5$ ps, rendering its predictions unreliable for further analysis. This observation reinforces our earlier conclusion that FNO struggles to learn the dynamics of nonlinear reactions occurring within localized high-temperature regions. This comparison further highlights that PARCv2 not only achieves better accuracy than popular state-of-the-art PIML models but also generalizes more effectively to out-of-distribution initial conditions.

\subsubsection{Low Velocity Extrapolation}\label{sec:results_low}
In \Cref{sec:inter}, we presented our finding that all three models exhibit deteriorated accuracy in lower-velocity cases. We now further investigate this failure mode when extrapolating beyond the coverage of the training set toward smaller velocities.

\begin{figure*}
    \includegraphics[width=\textwidth]{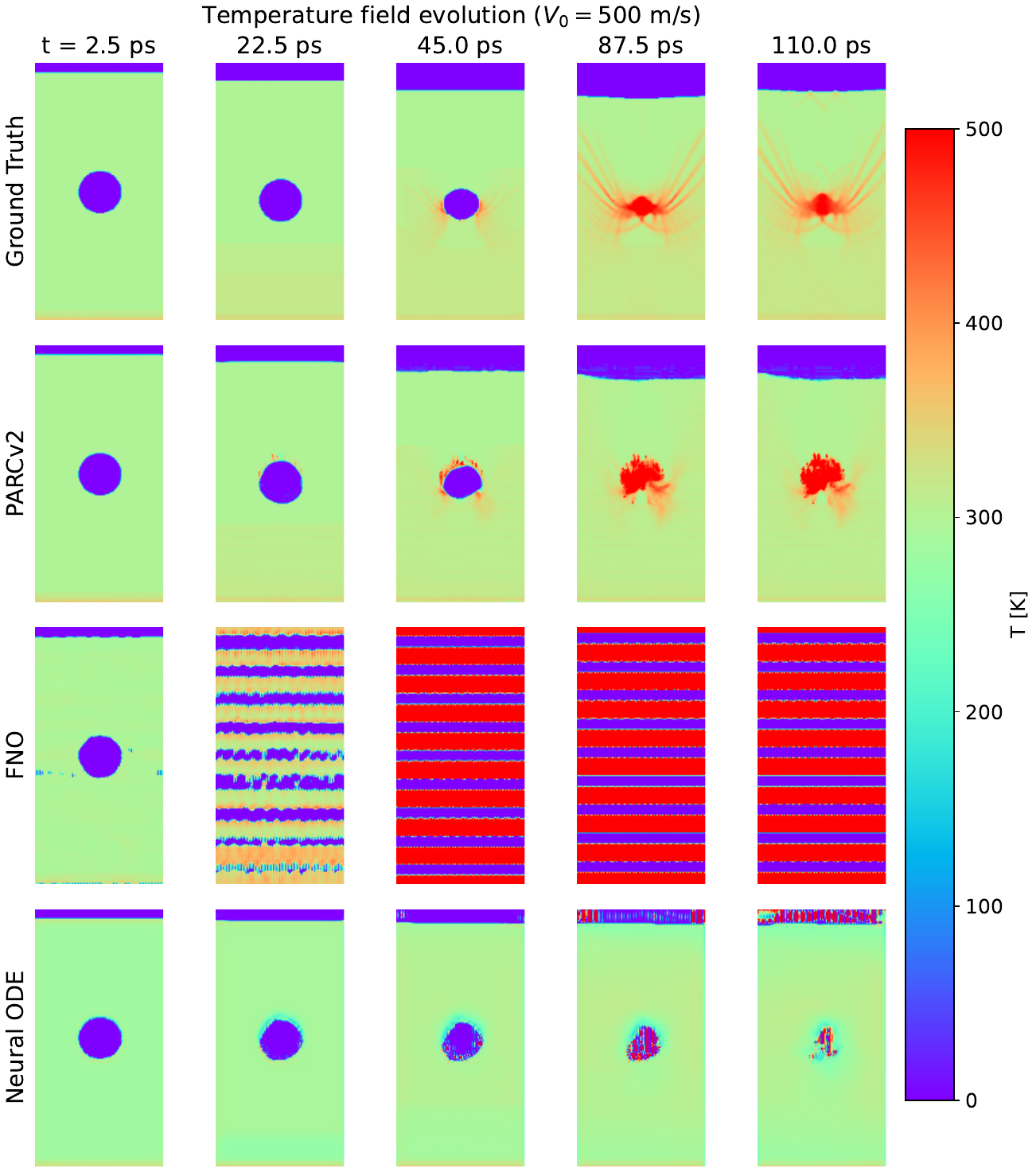}
    \caption{Temperature field evolution of initial velocity of 500 m/s.}\label{fig:temp_500}
\end{figure*}

We first observe that, while the RMSE indicates some decrease in performance for extrapolation to lower initial velocities—specifically, the error at $V_0=500$ m/s is approximately twice the lowest error achieved by the models, as shown in the top panel of \Cref{fig:mean_mae}—a more concerning issue arises when average temperature is considered. In this case, a nearly tenfold increase in relative error is observed. Notably, for the temperature, pressure, microstructure, and Y-velocity fields, the increase in RMSE during low-velocity extrapolation is much more pronounced for FNO (often exceeding a tenfold increase) compared to PARCv2 (typically around a twofold increase). Visual inspection of the predicted fields (e.g. \Cref{fig:temp_500} for temperature) reveals several numerical artifacts, including pore deformation, material interface warping, and anomalously high temperatures around the pore boundary.

Despite these issues, our observation of declining model performance at low velocities is consistent with the failure mode reported in \citet{cheng2024physics}, where PARCv2 was used to model supersonic flow around a cylinder. However, unlike that case, we do not observe numerically diverging values, persistent large-scale artifacts, or evidence that the failure is due to a shift in the underlying physical processes that are dominant in the system.

\subsection{Physics-driven Metrics}
\subsubsection{Distribution Functions of Temperature and Pressure}
\begin{figure}
    \subfloat[Temperature distribution function]{
        \includegraphics[width=\columnwidth]{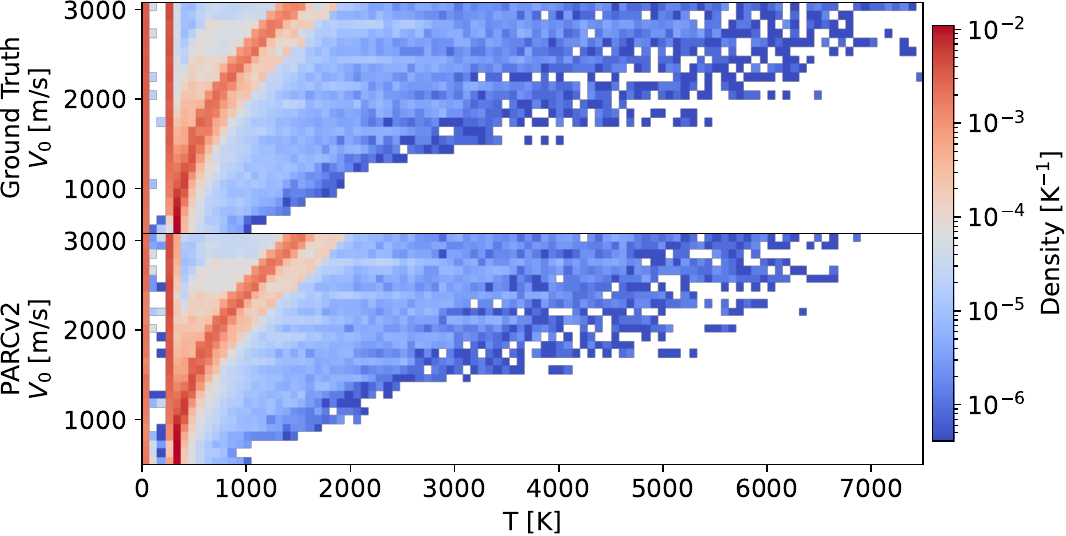}
    }\\
    \subfloat[Pressure distribution function]{
        \includegraphics[width=\columnwidth]{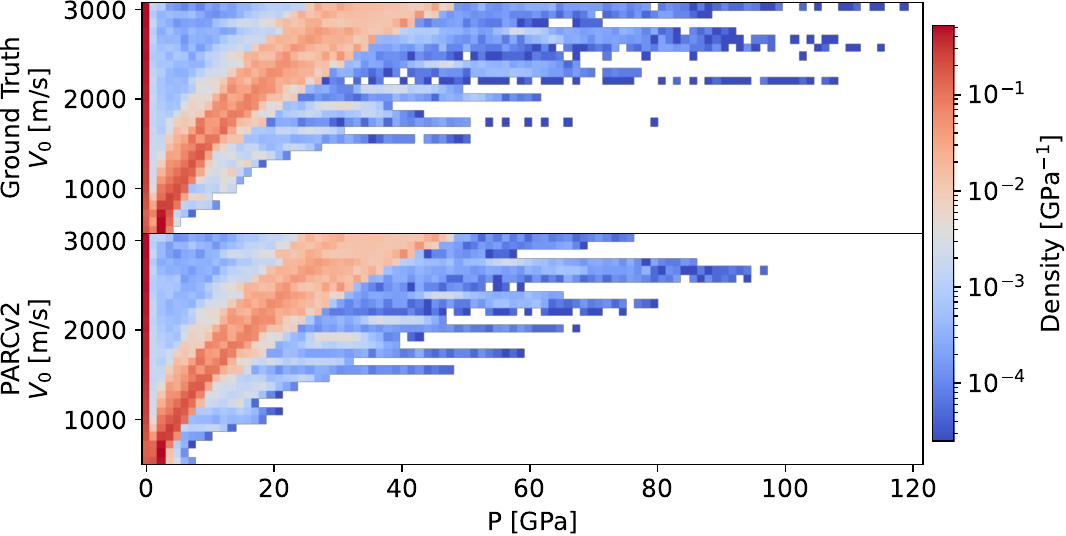}
    }
    \caption{Temperature (top) and pressure (bottom) distributions at the time of complete pore collapse for different initial velocity test cases.}\label{fig:temperature_v0_pdf}
\end{figure}

Following the approach of \citet{nguyen2024continuum}, we compared the distribution functions of the temperature and pressure fields at the time of complete pore collapse for each initial velocity, as shown in \Cref{fig:temperature_v0_pdf}, to evaluate the performance of PARCv2 on more physics-driven metrics. We observed that the model correctly captures the general trend of increasing peak temperature distribution with higher initial impact velocities. However, several systematic deviations are also apparent. First, while the ground truth distribution displays a clear gap between 0 K (representing vacuum at the top of the physical domain) and 200 K (the initial temperature of the material block), this gap is less prominent in the temperature distribution predicted by PARCv2. This suggests that the model struggles with discontinuous distributions. Additionally, the model consistently underpredicts the highest temperatures in the domain, as indicated by the blue region in the bottom panel of \Cref{fig:temperature_v0_pdf} not extending as far as in the top panel. A similar trend is observed in the pressure field: although the predicted pressure PDF closely matches the ground truth overall, PARCv2 consistently underpredicts the maximum pressure, with no predicted values exceeding 100 GPa.

\subsubsection{Shear Band Temperature Profile}
As the shear band phenomenon is of high interest in this work, we conducted a more focused study of the region exhibiting the strongest shear band signals. For each simulation, we selected the timestep at which shear bands were most prominent and compared the temperature profile along a vertical section at $X=29.30$ nm and $29.30<Y<175.79$ nm for several evenly distributed test cases, as shown in \Cref{fig:shear_band_profile_compare_0}. We found that PARCv2 consistently predicts the location and width of the most prominent shear bands, though the peak temperatures within the bands and their surrounding background are underestimated by 10–40 K. FNO performs well for the 1,800 m/s and 2,400 m/s cases, but it consistently misses more shear bands and exhibits larger RMSE values within the regions of interest.

This comparison confirms four key behaviors observed in our previous visual examinations: (a) PARCv2 outperforms FNO in predicting shear band formation and evolution; (b) all models achieve their best prediction accuracy within the 1,200 m/s to 2,400 m/s range; (c) weaker shear bands are often missed by all models, including PARCv2; and (d) predictions from the ResNet-based neural ODE are unsatisfactory.

We further analyzed the temperature profile within the region of interest for the high-velocity extrapolation test case, as shown in the bottom panel of \Cref{fig:shear_band_profile_compare_1}. Notably, the ground truth sequence indicates a general absence of strong shear band effects in higher-velocity cases. Unlike other cases within the training set, where the ground truth (solid lines) exhibits frequent but weak temperature variations, the $V_0=3000$ m/s case features one large, wide drop in temperature. PARCv2 accurately captures this feature, although it underpredicts the peak temperature. Nevertheless, compared to FNO, the PARCv2 prediction yields an RMSE value approximately 50\% lower and is free of any nonphysical increases in the temperature profile, such as a peak in temperature at $y\sim 110$ nm in the FNO prediction when $V_0$ = 3000 m/s in the bottom panel of \Cref{fig:shear_band_profile_compare_1}.

A lack of accurately predicted shear bands at low velocity is also evident from the top panel of \Cref{fig:shear_band_profile_compare_0}. While PARCv2 predicts some temperature increase at the expected locations of shear bands, the predicted peak temperatures show the greatest deviation among all cases, and the background continuum is poorly captured. The predicted continuum is consistently lower than the ground truth, peak temperatures in the shear bands are more than 50 K lower, and the predicted bands are significantly wider than those in the ground truth.

\begin{figure*}
    \includegraphics[width=\textwidth]{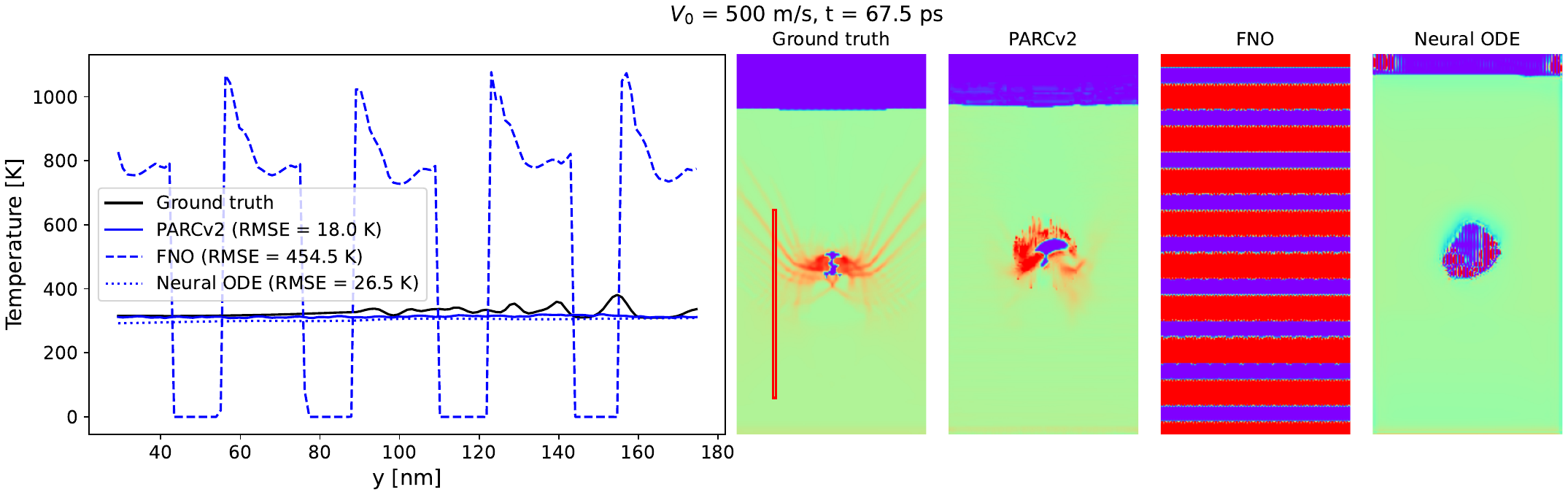}
    \includegraphics[width=\textwidth]{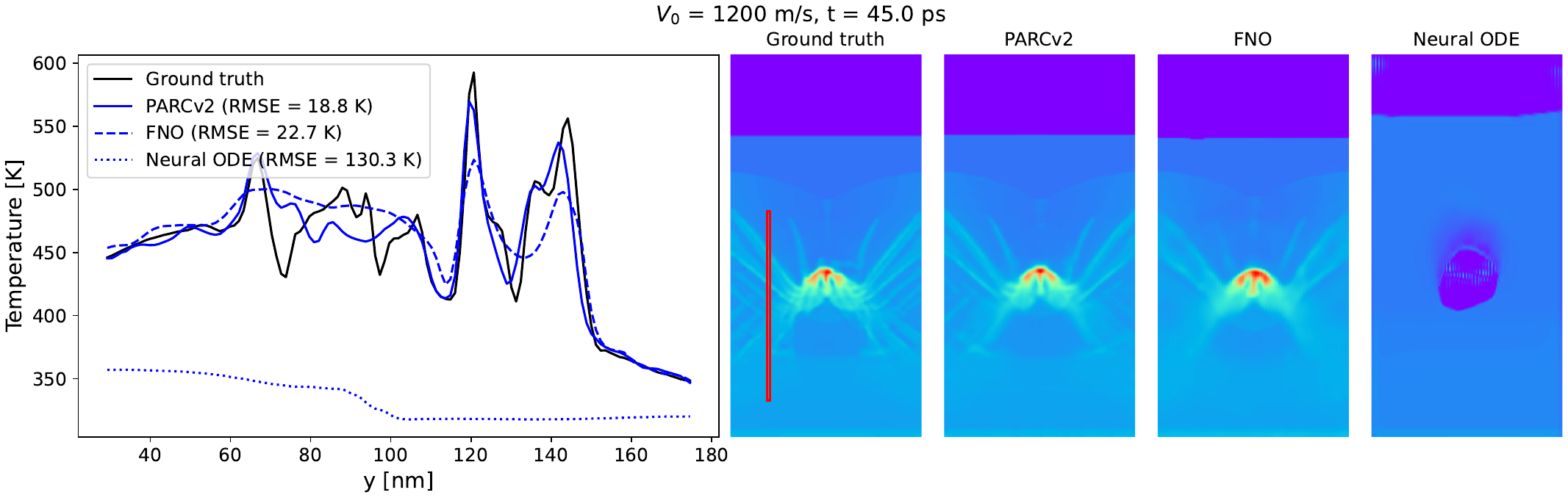}
    \includegraphics[width=\textwidth]{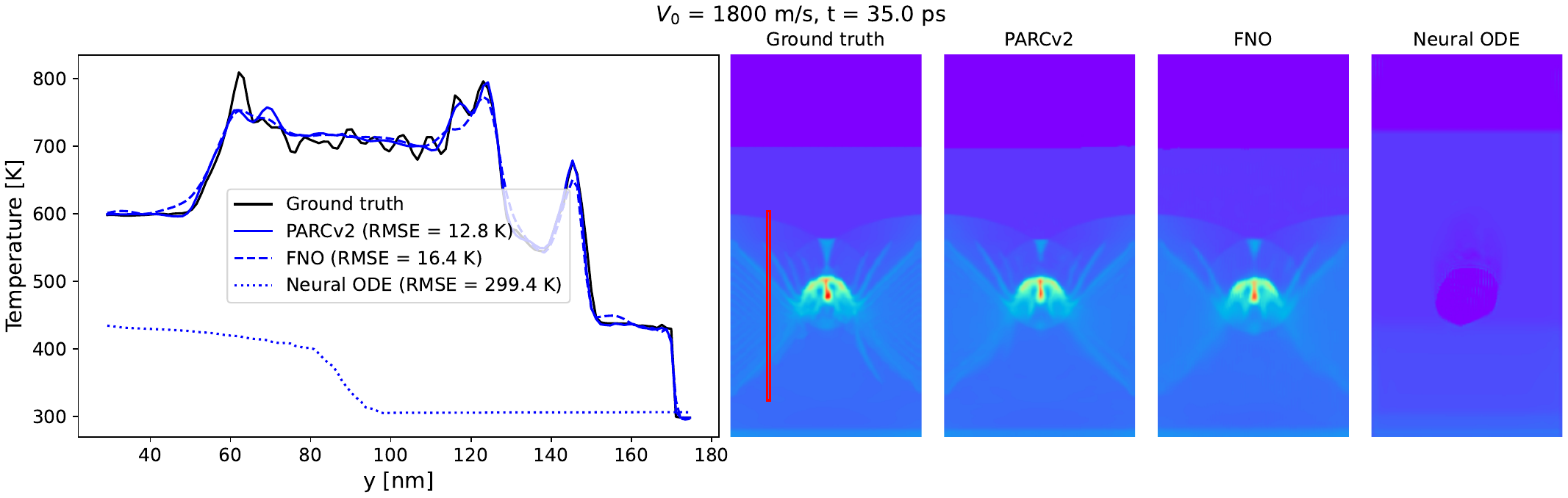}
    \caption{Shear band profile in temperature for initial impact velocity $V_0$=500, 1200, 1800, 2400 and 3000 m/s. We examine the temperature profile alone a vertical slice inside the red rectangle. Snapshots at the time of complete pore collapse, when the shear bands is most significant, are chosen to be examined.}\label{fig:shear_band_profile_compare_0}
\end{figure*}
\begin{figure*}
    \includegraphics[width=\textwidth]{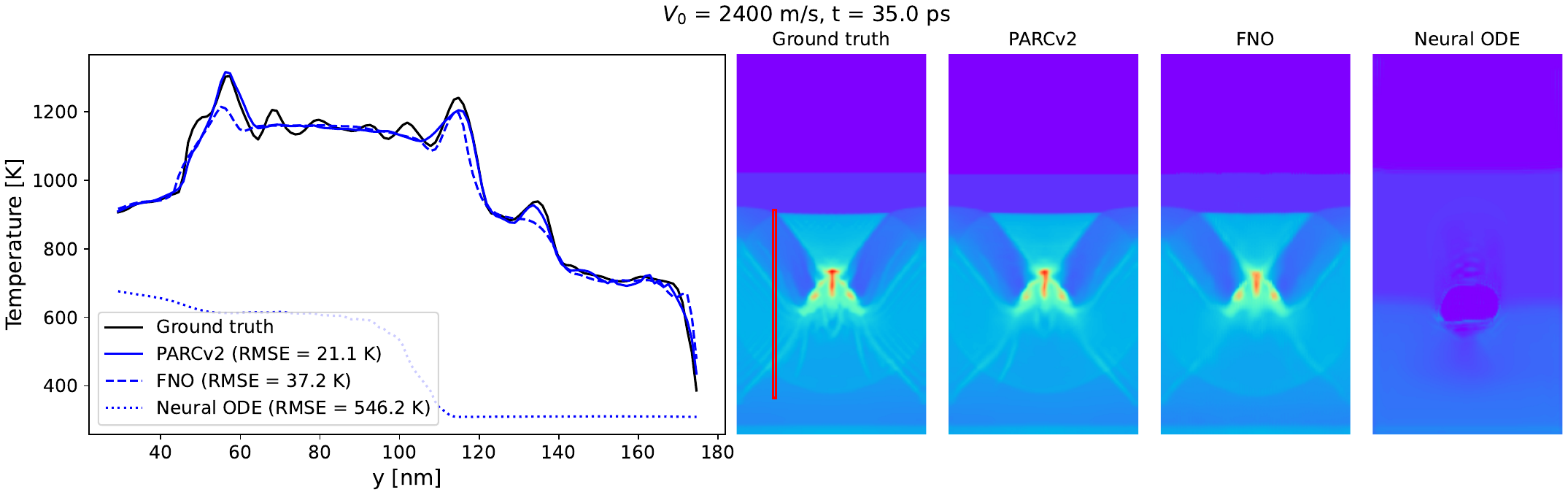}
    \includegraphics[width=\textwidth]{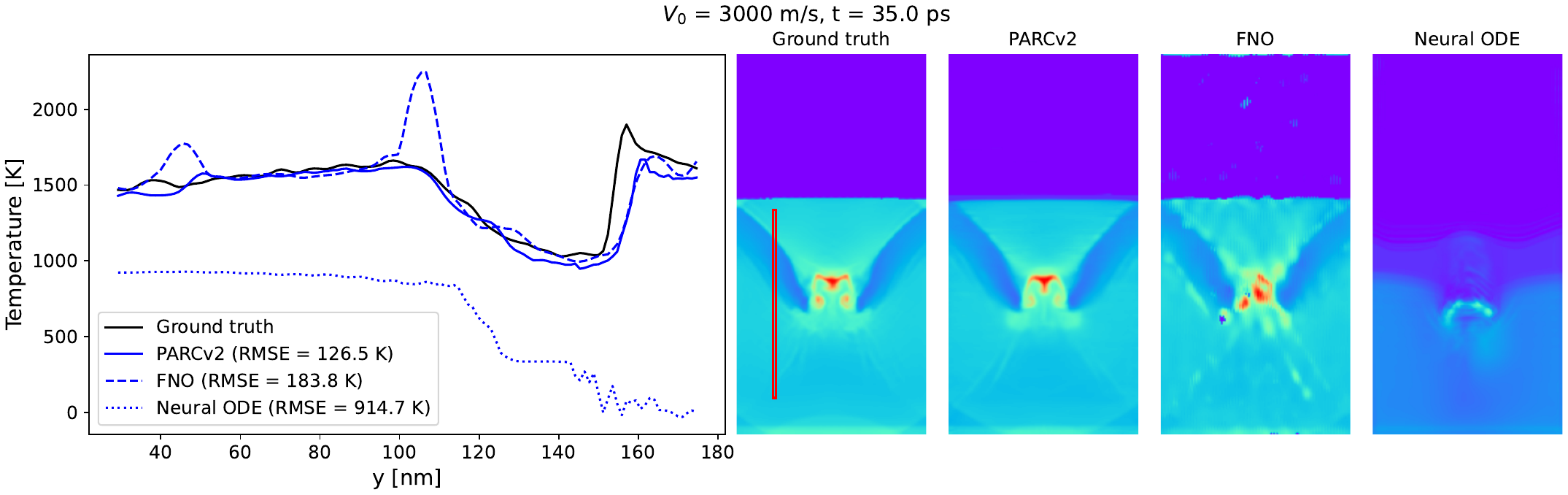}
    \caption{Continuation of \Cref{fig:shear_band_profile_compare_0}}\label{fig:shear_band_profile_compare_1}
\end{figure*}

\subsubsection{Dominant Shear Band Location, Width and Temperature Increase}
We further examined the prediction quality of the dominant shear band by extracting its location, width, and temperature increase from \Cref{fig:shear_band_profile_compare_0} and \Cref{fig:shear_band_profile_compare_1}, with results summarized in \Cref{tab:dominant_shear_band_physics}. Models achieving the smallest deviation from ground truth in each physics-driven metric and impact velocity are underscored in the corresponding column. For $V_0 = 1800$ m/s and $V_0 = 2400$ m/s, both PARCv2 and FNO predicted the peak-temperature location with zero error and reproduced the dominant shear band width with only a one-pixel deviation from ground truth. However, in terms of temperature increase relative to the surrounding continuum, both models underpredict compared to ground truth—indicating a systematic tendency to miss higher-frequency variations, a telltale sign of spectral bias. Further inspection of the $V_0 = 2400$ m/s case in the top panel of \Cref{fig:shear_band_profile_compare_1} shows that FNO (dashed line) consistently underpredicts the temperature near strong shear bands. Although this effect is not captured by the physics-driven metrics in \Cref{tab:dominant_shear_band_physics}, the elevated RMSE values correctly reflect the evident deviation from ground truth.
\begin{table}
    \centering
    \begin{tabular}{cc|c@{\hskip 0.2in}c@{\hskip 0.2in}c@{\hskip 0.2in}c}
    \hline
    $V_0$ [m/s] & & Ground Truth & Improved PARCv2 & FNO & Neural ODE \\
    \hline
    & $y_{peakT}$ [nm] & 154.6908 & & & \\
    500 & $\Delta y$ [nm] & 15.2348 & Fail & Fail & Fail \\
    & $\Delta T$ [K] & 70.4020 & & & \\
    \hline
    & $y_{peakT}$ [nm] & 144.1437 & 141.7999 & \underline{142.9718} & \\
    1200 & $\Delta y$ [nm] & 18.1644 & \underline{18.1645} & 14.6487 & Fail \\
    & $\Delta T$ [K] & 155.1000 & \underline{111.9133} & 52.05890 & \\
    \hline
    & $y_{peakT}$ [nm] & 145.3156 & \underline{145.3156} & \underline{145.3156} & \\
    1800 & $\Delta y$ [nm] & 11.1330 & \underline{9.96115} & \underline{9.96115} & Fail \\
    & $\Delta T$ [K] & 132.975 & \underline{129.8824} & 107.6409 & \\
    \hline
    & $y_{peakT}$ [nm] & 114.8462 & \underline{114.8462} & \underline{114.8462} & \\
    2400 & $\Delta y$ [nm] & 11.1330 & \underline{13.4768} & \underline{13.4768} & Fail \\
    & $\Delta T$ [K] & 140.7800 & 90.4980 & \underline{111.3630} & \\
    \hline
    \end{tabular}
    \caption{Dominant shear band location $y_{peakT}$, width $\Delta y$ and temperature increase $\Delta T$ at $X=29.30$ nm for a number of impact velocities. Model with the prediction closest to ground truth are underline for each metrics, and models failed to predict shear band patterns are labeled with ``Fail" in the corresponding column. $V_0$ = 3000 m/s was skipped due to the lack of shear band phenomenon.}
    \label{tab:dominant_shear_band_physics}
\end{table}

For lower impact velocity cases, PARCv2 again demonstrates superior accuracy compared to FNO. At $V_0 = 1200$ m/s, PARCv2 outperforms in predicting both the width and temperature increase of the dominant shear band in \Cref{tab:dominant_shear_band_physics}, while trailing slightly behind FNO in the predicted location. However, this simple descriptive metric is misleading: as shown in the middle panel of \Cref{fig:shear_band_profile_compare_0}, the dominant shear band exhibits a double-peak structure that FNO completely misses. Although PARCv2 shows some misalignment in the higher-temperature peak, it correctly captures the existence of the double-peak structure. Consistent with this observation, the elevated RMSE values for FNO reflect its larger deviation from ground truth.

Similar to previous cases, the $V_0 = 500$ m/s impact velocity proves challenging for all three models, as none of the predictions exhibit a clear shear-band pattern. Consequently, all three models are labeled as Failed in \Cref{tab:dominant_shear_band_physics}. The $V_0 = 3000$ m/s case cannot be assessed using these metrics, since the shear-band patterns are statistically indistinguishable from temperature oscillations present in the continuum, and is therefore omitted from the table.

\section{Discussion}\label{sec:discussion}
While PARCv2 demonstrates the best overall performance among popular PIML models and achieves reasonable accuracy for many dynamical processes --- including the formation and evolution of dominant shear bands and hotspots, shocks generated by the reverse ballistic impacts, and distortion of material interfaces --- we also observed several limitations of current PIML models, which we address in more detail below.

\subsection{Spectral Bias}\label{sec:discussion_details}
\begin{figure*}
    \centering
    \includegraphics[width=\textwidth]{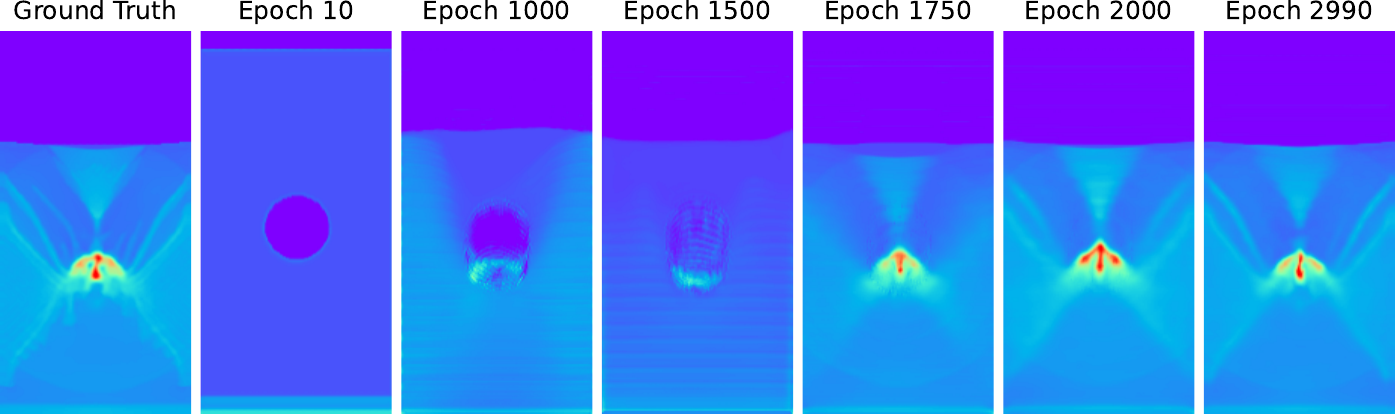}
    \caption{16-step roll-out prediction at various epoches during training. We conclude that the model tends to learn the low frequency dynamics first and finer details later, a telltale sign of spectral bias.}
    \label{fig:training_sequence}
\end{figure*}

As observed in previous sections, the predicted fields display diffused shock fronts and shear bands, leading the model to miss some of the smaller and weaker shear bands. This limitation can be attributed to the low-frequency bias of neural networks (e.g., \citet{xu2024overview}). In our problem, this bias manifests in the model learning large-scale patterns (e.g., regions of high temperature, shock waves, reverse ballistics) earlier in training, while smaller-scale patterns (e.g., shear bands, shape and asymmetry of the high-temperature regions) are learned later. We examined our model at various training epochs and confirmed this hypothesis in \Cref{fig:training_sequence}. The evolution of the material interface and the motion of the high-temperature region were learned within the first 1,500 epochs. The ignition of the high-temperature region began to appear with reasonable detail around epoch 1,750, and the shear band patterns became predictable after 2,000 training epochs.

\begin{figure*}
    \centering
    \includegraphics[width=0.629\linewidth]{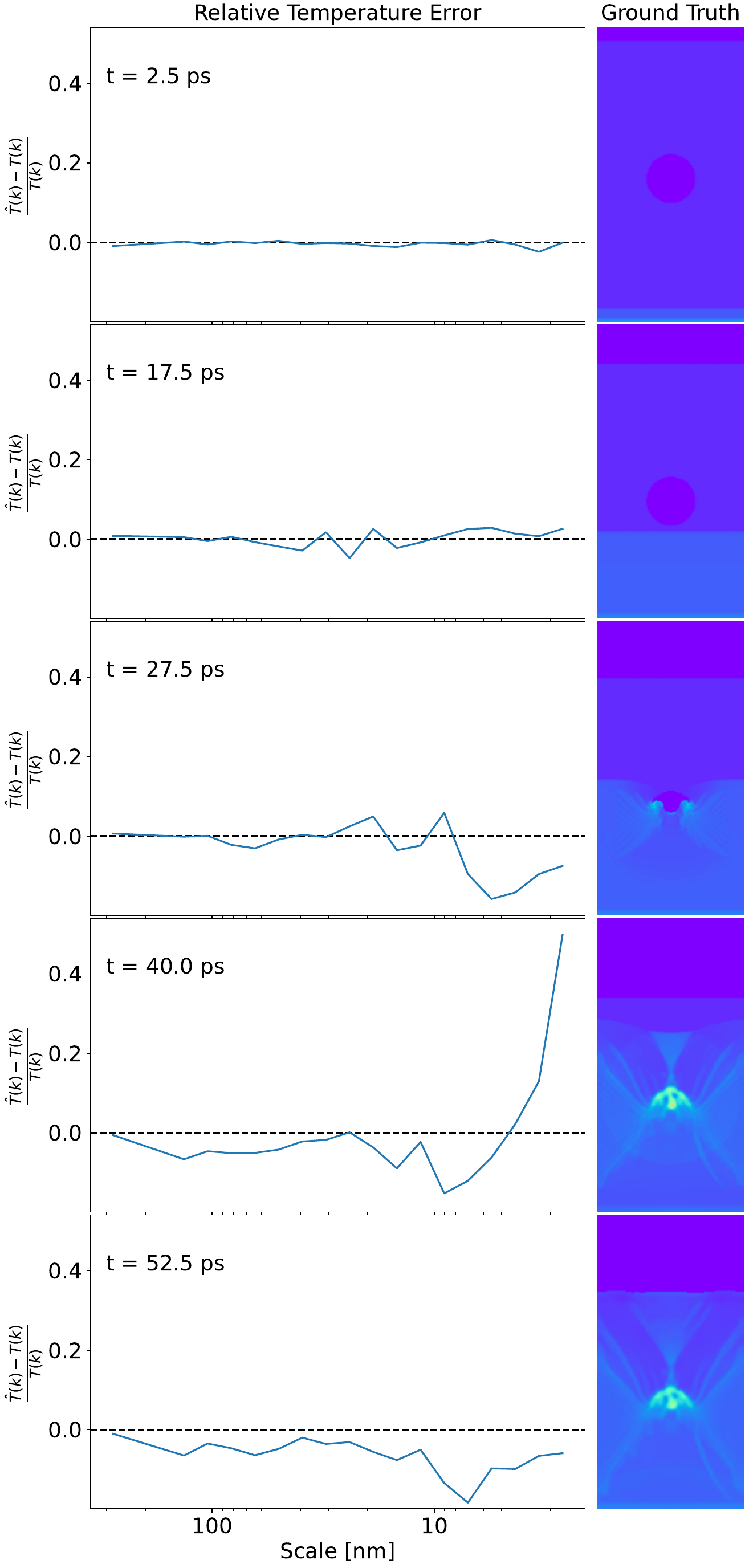}
    \caption{Relative error of temperature spectrum at various time for $V_0$ = 1800 m/s.}
    \label{fig:spectrum_compare}
\end{figure*}

To further investigate the spectral bias of PARCv2, we present the relative error of the temperature spectrum as a function of time in \Cref{fig:spectrum_compare}. As previously discussed, blurry predictions and missing weaker shear bands persist across impact velocities; therefore, we focus on the case with $V_0 = 1800$ m/s. At the start of the roll-out, shown in the first two panels of \Cref{fig:spectrum_compare}, the relative prediction error remains approximately constant across spatial scales. However, at $t = 27.5$ ps, during pore collapse, a sharp increase in error appears at the high-frequency end (corresponding to smaller spatial scales), as seen in the third panel. When shear band patterns become most prominent at $t = 40$ ps, the deviation for temperature structures smaller than $10 nm$ rapidly increase, reaching 50\% at the smallest spatial scale resolvable by the simulation. From this snapshot alone, it is unclear whether the elevated high-frequency error originates from inaccuracies in predicting weaker, non-dominant shear bands or from the fine-scale hotspot geometry, as both features exist below the $10 nm$ scale. Examination of the final snapshot, however, rules out the hotspot geometry as the primary source: the hotspot shape remains largely unchanged over time, while the error drops significantly as shear-band-induced temperature rises dissipate. Together, these results provide evidence that the misprediction of weaker shear bands is primarily due to the spectral bias of the machine learning model.

\begin{figure}
    \centering
    \includegraphics[width=0.9\columnwidth]{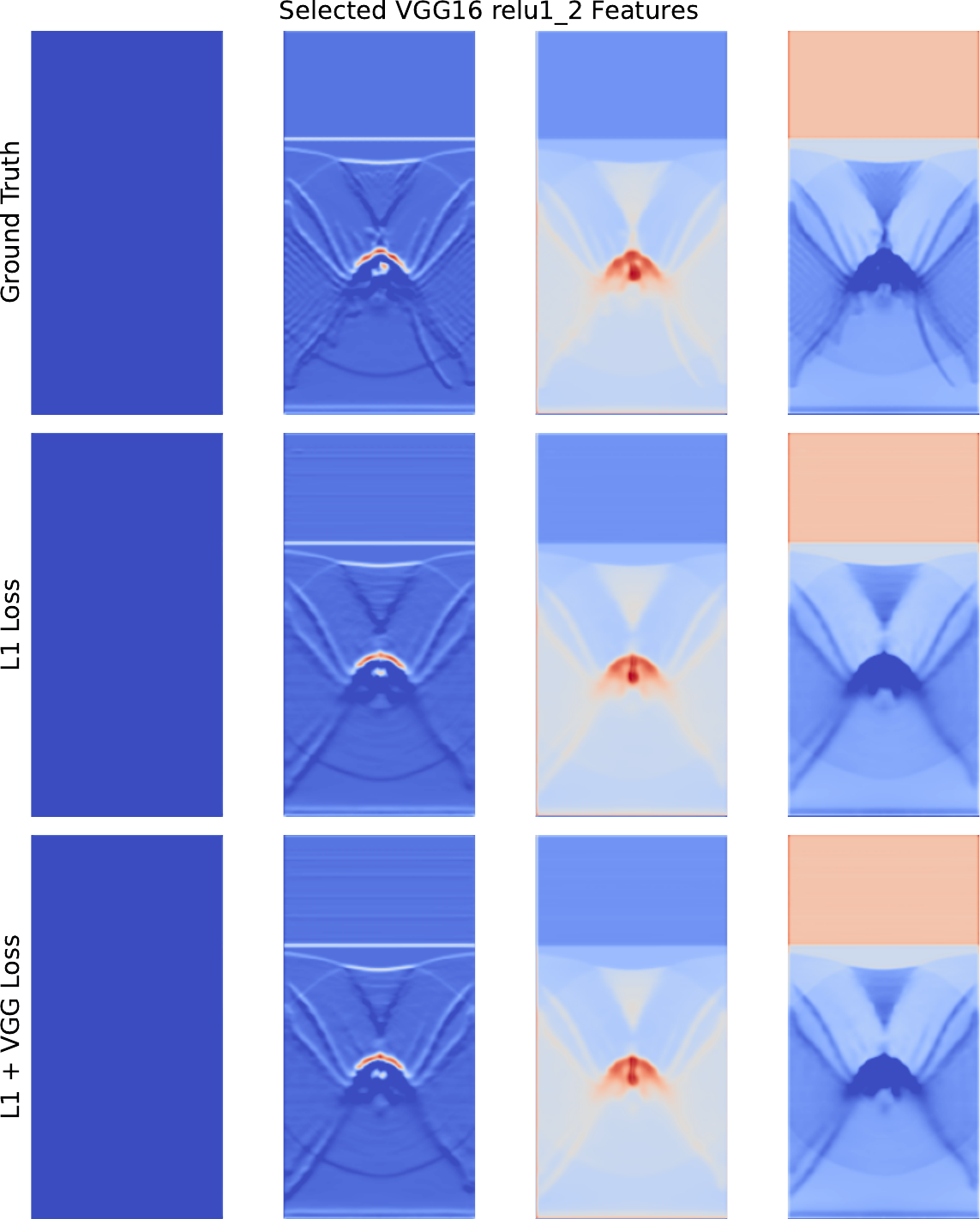}
    \caption{Feature extraction with pretrained VGG model on ground truth simulation, prediction from model trained with L1 data loss only and prediction from model further finetuned with perception loss. The pretrained VGG16 model can indeed extract the shear band features in some channels but not others. Nevertheless, this indicates that these features can indeed be utilized as perception loss.}
    \label{fig:vgg_patterns}
\end{figure}

\begin{figure}
    \centering
    \includegraphics[width=\columnwidth]{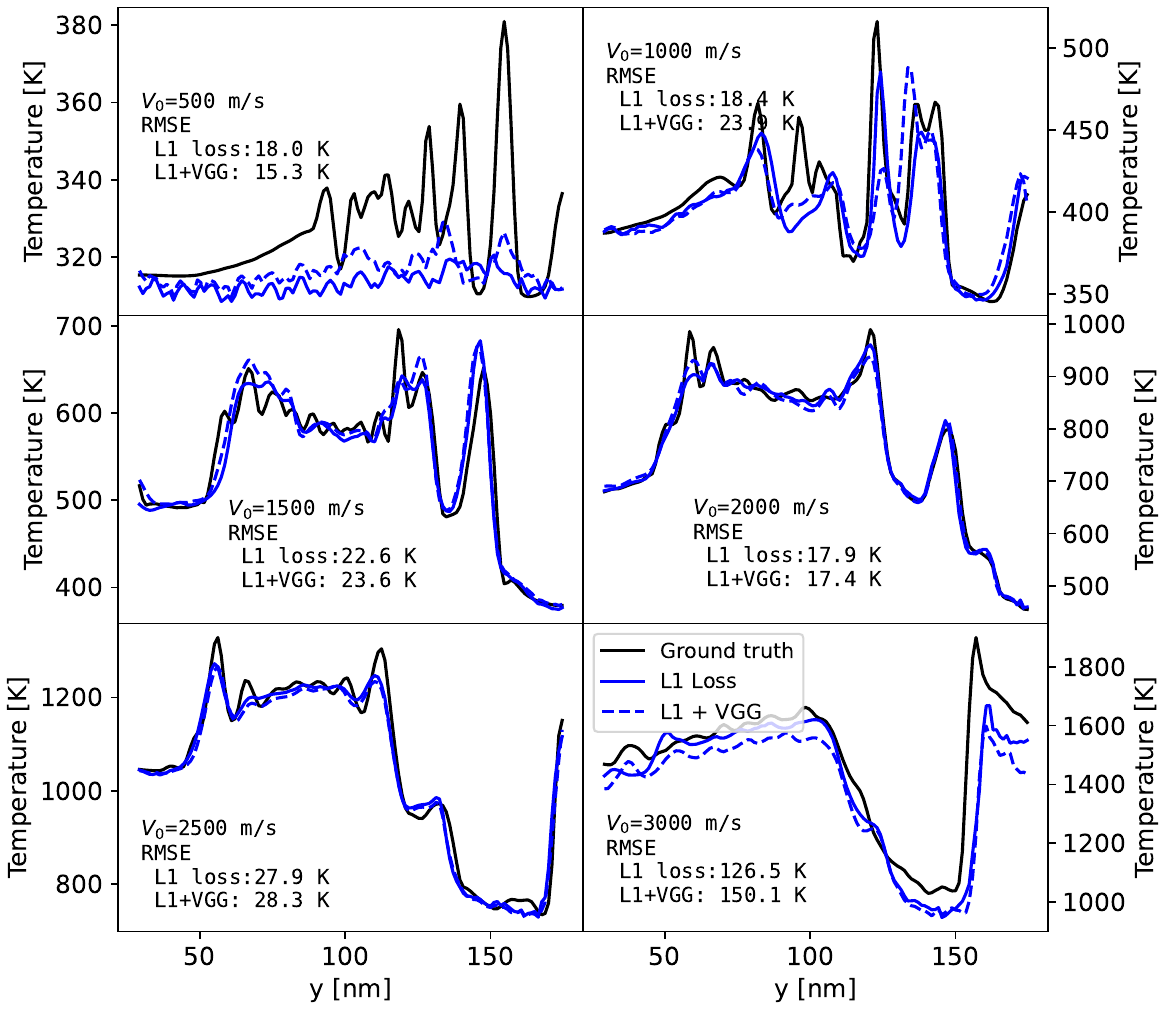}
    \caption{Comparison of shear band profile with model trained on L1 data loss only, finetuned with VGG loss and finetuned with wavelet loss. We did not observe any meaningful increases in performance in the region of scientific interests.}
    \label{fig:shear_band_profile_compare}
\end{figure}

To address this issue and improve prediction accuracy, we turned to progress made in the computer vision community. It has been suggested that using outputs from shallow layers of a pretrained network in image synthesis and super-resolution tasks can significantly enhance image sharpness \citep{johnson2016perceptual}. Given the distinct nature of image classification and physics simulation, we first examined features extracted by an ImageNet \citep{deng2009imagenet} pretrained VGG-16 classifier \citep{simonyan2014very} to ensure that patterns of interest could be identified using this pretrained model. We selected the \textit{relu1\_2} layer features for use as additional perceptual loss terms, since shear band patterns were clearly extracted at this specific network depth. Some shear band structures appeared in the extracted features from this layer, as shown in \Cref{fig:vgg_patterns}. We then fine-tuned our network with this added perceptual loss term for an additional 1,000 epochs using a learning rate of $10^{-5}$. While we observed a non-trivial decrease in both data and perceptual loss, we did not see noticeable improvements in the predictions. Visual inspection of the rollout sequence revealed no significant changes, even when comparing the VGG-extracted features in the second and third rows of \Cref{fig:vgg_patterns}. Additionally, improvements in temperature prediction for shear bands in \Cref{fig:shear_band_profile_compare} were minimal. In terms of mean absolute difference from the ground truth (indicated in each panel), we found that the fine-tuned model actually performed worse in both high- and low-speed extrapolation cases.

One major drawback of the VGG perceptual loss is the increased training time, as it requires backpropagation through part of another deep learning model. Given that our model already demands substantial computational effort to train, we explored an alternative approach by using discrete wavelet transforms to extract high-frequency features, which we then used as an additional perceptual loss term. We selected the Haar wavelet and constructed the wavelet perceptual loss from the highest-frequency wavelet coefficients. We repeated the fine-tuning process described above, substituting the VGG perceptual loss with the wavelet perceptual loss. However, we again observed no improvement in the shear band temperature profiles.

Concerned that the added loss term may have drastically altered the loss landscape, making the previous L1-only trained model poorly situated in the new space, we retrained the model from scratch using the wavelet perceptual loss. Although this did not significantly improve prediction performance, we observed a drastically faster convergence, with the model producing usable results within just 900 epochs. We therefore conclude that while perceptual loss does not enhance the sharpness of predictions in our problem, its emphasis on high-frequency components is nonetheless beneficial for accelerating training. We hypothesize that this acceleration could extend to many other mesh-based PIML models across a range of physical systems.

\citet{ng2024spectrum} proposed that using a generalized Lp norm with large $p$ values could improve learning of high-frequency components and increase predictive accuracy. In theory, reducing per-step error should yield improved rollout performance. To test this, we fine-tuned PARCv2 using $p = 10$ for 1000 epochs with a constant learning rate of $10^{-6}$. However, our results show no benefit for our problem. Despite a significant decrease in Lp loss values on both training and validation sets, the rollout predictions exhibited degraded performance, including numerical artifacts and signs of instability. Many of these artifacts originated from dragging of the circular pore and horizontal banding in regions of constant field values. Compared to the much smoother field and stationary setup in \citet{ng2024spectrum}, we conclude that this method is less effective for time-recursive PIML models dealing with complex, dynamic physics.

Our investigation into improving prediction sharpness highlights that, in the context of physics-informed machine learning, addressing low-frequency bias remains an open challenge. Although several techniques have been proposed for deep neural networks—such as phase shifting \citep{cai2020phase}, Fourier features \citep{tancik2020fourier}, and multi-stage networks with sinusoidal activation functions \citep{wang2024multi}—these methods all explicitly rely on spatial coordinates. This dependency can compromise shift invariance and limit spatial extrapolation capabilities. Moreover, it has been shown that emphasizing high-frequency features can impair a model's generalization ability despite reducing prediction error \citep{abello2021dissecting}. While DNNs are not generally expected to generalize across different initial conditions, this loss of generalizability (a) undermines a key strength of mesh-based PIML models like ours, which can infer across a range of initial conditions with reasonable accuracy, and (b) introduces the risk of numerical instability due to accumulated errors in temporally recursive models.

We hypothesize that addressing the sharpness issue effectively will require entirely new neural network architectures. Most modern architectures rely on hierarchical information summarization: shallow layers learn local features, which are then aggregated by deeper layers to capture global dynamics. We believe this structure inherently deprioritizes the accurate modeling of fine-scale features, as training gradients initially flow to layers responsible for large-scale dynamics. A promising solution may lie in designing architectures that can learn both large and small scales at the same network depth. Such a design could help resolve the issue of missing smaller and weaker shear bands and enhance the overall contrast of predicted fields.

\subsection{Enforcing Boundary Conditions for Mesh-based PIML Models}
While numerical solvers can handle all types of boundary conditions with minimal user intervention, this is not the case in AI/ML based simulations for physics applications. Physics-naive models generally apply no special treatment to boundary conditions—unsurprising, given that such considerations are of limited importance in the computer vision community. In the realm of physics-informed machine learning (PIML), however, enforcing boundary conditions depends heavily on model architecture, and a variety of methods have been proposed.

The most straightforward approach involves placing collocation points along the boundary and applying a soft constraint via additional loss terms. This method is the most commonly used for meshless PIML models (e.g., \citet{raissi2019physics,berg2018unified,sirignano2018dgm}). Other strategies include using variational principles \citep{yu2018deep,han2018solving} or Petrov-Galerkin domain decomposition \citep{kharazmi2019variational,kharazmi2021hp} to impose boundary conditions indirectly. However, multiple studies (e.g., \citet{lu2021deepxde,chen2020comparison,wang2020understanding}) have shown that the inexact nature of these methods can hinder convergence and reduce prediction accuracy, especially in problems with complex geometries. To address this, recent efforts have focused on exactly enforcing boundary conditions (e.g., \citet{lyu2020enforcing,sukumar2022exact}). Unfortunately, these exact methods often require explicit coordinate dependencies, making them difficult to translate to mesh-based models such as convolutional or graph neural networks.

For mesh-based PIML models, custom solutions have emerged. \citet{ren2022phycrnet} introduced a practical and user-configurable strategy for enforcing boundary conditions using built-in padding modes from popular deep learning frameworks. In their method, Dirichlet boundary conditions (e.g., zero pressure at an outlet) are enforced via constant padding, Neumann conditions (e.g., zero velocity gradient) via reflective padding, and periodic boundaries via circular padding. We adopted this approach in our model. For graph neural networks, \citet{pfaff2020learning} proposed marking boundary nodes with a non-updating attribute, effectively freezing their values to enforce Dirichlet conditions. However, extending this to Neumann boundaries is significantly more difficult.

In our work, we apply zero-gradient (Neumann) boundary conditions across all fields following \citet{ren2022phycrnet}. However, we encountered several practical caveats. While the method is easy to implement, most deep learning libraries apply the same padding rule to all boundaries. Enforcing distinct boundary conditions (e.g., fixed inlet velocity and zero-gradient outlet) requires low-level customization of padding before each convolution. Moreover, constant padding for Dirichlet boundaries is straightforward at the input layer but becomes problematic in deeper layers. For instance, padding the inlet with a fixed velocity value works at the first layer, but the corresponding values after subsequent layers are no longer the original inlet velocity — they are transformed by layer weights. This implies that the padded values must be updated dynamically after each weight update to remain consistent, which is not supported natively by any existing deep learning code frameworks.

Enforcing Neumann boundary conditions also poses challenges. Reflect padding enforces zero gradients only under symmetric schemes like central differencing. For asymmetric schemes such as upwinding, zero-gradient conditions instead require constant padding. However, because the specific behavior of each convolution filter is unknown ahead of time, modifying padding based on filter orientation becomes infeasible. Therefore, robust and generalizable methods for \textit{hard} enforcement of boundary conditions in mesh-based models remain an important area for further research and development.

\subsection{Prediction of Low Impact Velocity/ Low shock strength cases}
In \Cref{sec:results_low}, we highlighted a puzzling behavior of our deep learning model: extrapolations to lower initial velocities result in significantly higher loss values and a failure to capture the dynamics. This is not the first observation of such behavior. While similar trends have been reported in \citet{cheng2024physics} --- where reduced accuracy at lower velocities was attributed to the complex transition from subsonic to supersonic wake flow --- no such change in dynamics is present in our problem. In fact, \citet{weger2025shear} showed that a power law can describe the spacing of dominant shear bands across the entire initial impact velocity range, suggesting that the underlying physical processes remain consistent.

We note that lower initial impact velocities correspond to smaller magnitudes of physical quantities (e.g., temperature, pressure, velocity), which we hypothesize is a key contributor to the model's degraded performance. As discussed in \Cref{fig:training_sequence}, smaller variations in data are typically learned at later stages of training. While such deviations are negligible at higher impact velocities, they can be comparable in scale to the temperature contrast between shear bands and their surroundings at lower velocities. Consequently, the model's inability to resolve these small differences may lead to an overall failure to learn the dynamics in these cases.

Although the universal approximation theorem assures that deep learning models can represent any function in theory, in practice, prediction errors often plateau even with larger model sizes and prolonged training, as shown in \citet{wang2024multi}. This can be attributed to the bias–variance tradeoff, which limits the model’s generalization capability beyond its training distribution. We believe that normalizing data through a learnable normalization flow, parameterized by initial conditions or physical time, may help mitigate this failure mode. We are currently exploring such methods.

A second likely cause is the distribution shift associated with different initial conditions. As with any machine learning application, model performance depends critically on the similarity between the training and inference data distributions. While this assumption typically holds in computer vision tasks, it is violated in the modeling of shear band formation. Different impact velocities produce vastly different distributions in temperature, pressure, and velocity. For example, maximum temperatures exceed 10,000 K at high velocities, while they barely reach 500 K at the lowest. This wide variation adds further complexity and exacerbates the learning challenge.

Finally, the mathematical structure of our model --- based on learning time derivatives --- may further worsen the issue. Lower velocity impact cases exhibit slower dynamics, which reduces the signal-to-noise ratio for learning time evolution and makes the model more susceptible to this specific failure mode.

\section{Conclusion}\label{sec:conclusion}
An improved implementation of physics-aware recurrent convolution algorithm called PARCv2 was employed to predict the temperature, pressure and interfacial dynamics in a shock-induced pore collapse problem. The physics of this problem is relevant to hotspot formation and growth in energetic materials, which is the initiating feature leading from shock to detonation. We show that the PARCv2 algorithm provides superior predictions when compared to two other popular PIML approaches, namely FNO and neural ODE. When tested for cases within the range of the training dataset PARCv2 consistently provides predictions in close agreement with ground truth.

The improved PARCv2 model achieves state-of-the-art performance in modeling pore collapse in solid materials under weak-to-moderate shock regimes, capturing most physical processes with satisfactory accuracy across a range of initial conditions. However, persistent issues remain --- namely, blurred predictions, missing weaker shear bands, and degraded performance at lower impact velocities --- which affect all PIML models tested. Although the computer vision community has made significant progress on similar challenges, our attempts to incorporate their techniques --- such as perceptual losses and high-frequency feature learning --- into physics-informed machine learning architectures and training strategies yielded only limited improvements. These methods failed to fully resolve the observed failure modes.

We conclude that such limitations can hinder the broader adoption of deep learning models in complex computational fluid dynamics and solid mechanics applications, despite the substantial reductions in simulation time and computational cost they offer. Deeper investigation into these failure modes is essential for advancing PIML model development and will have broad impact on the computational physics community.

\begin{acknowledgments}
This work was supported by the Army Research Office (ARO) Energetics Basic Research Center (EBRC) program under Grant No. W911NF-22-2-0164.
\end{acknowledgments}

\section*{Conflict of Interest}
The authors have no conflicts to disclose.

\section*{Data Availability}
The data that support the findings of this study are available from the corresponding author upon reasonable request.

\bibliography{aiptemplate}

\clearpage

\appendix
\section{Continuum Governing Equations and Constitutive Laws}
\subsection{Continuum Model}

The continuum-scale conservation laws and material models are as follows. The mass, momentum, and energy equations are cast in Eulerian form \cite{rai2015framework,rai2015mesoscale}:
\begin{equation}
    \frac{\partial \rho}{\partial t} + \nabla .(\rho \mathbf{u}) =0
\end{equation}

\begin{equation}
\frac{\partial (\rho \mathbf{u})}{\partial t} +\nabla . (\rho \mathbf{u} \otimes \mathbf{u} + p \mathbf{I}) = \nabla. \mathbf{S}
\end{equation}
\begin{equation}
\frac{\partial \rho \left( e+ \frac{1}{2} \mathbf{u}.\mathbf{u}\right)}{\partial t} + \nabla. \left(\rho \left(e+\frac{1}{2} \mathbf{u}.\mathbf{u}\right)+p\right)\mathbf{u} = \nabla. (\mathbf{S}.\mathbf{u})  -\nabla. 
\mathbf{q}
\label{eq:E}
\end{equation}

Here, $\rho$ is the material density; $\mathbf{u}$ is the velocity vector; $e$ is the specific internal energy; $T$ is the temperature; $p$ and $\mathbf{S}$ are respectively the volumetric and the deviatoric components of the Cauchy stress tensor: $\sigma = -p \mathbf{I} + \mathbf{S}$  where $\mathbf{I}$ is the identity tensor. The heat flux $\mathbf{q}$ is given by Fourier's law $\mathbf{q}=-\chi \nabla T$, where $\chi$ is the thermal conductivity.

The above equations are supplemented with constitutive models to close the system, as follows.

\subsection{Stress evolution equation}
The deviatoric stress tensor S is obtained from the following hypo-elastic constitutive model in a rate formulation (Prandtl-Reuss \cite{khan1995plasticity, huang2021explicit}):

\begin{equation}
\frac{\partial \rho \mathbf{S}}{\partial t} +\nabla .(\rho \mathbf{S} \otimes \mathbf{u}) + \rho(\mathbf{S}\mathbf{W}-\mathbf{W}\mathbf{S}) = 2\rho G(\mathbf{D}^{'}-\mathbf{D}_{pl}
)
\label{eq:S_evolution}
\end{equation}

where the left hand side is the objective Jaumann rate for the evolution of the deviatoric stress $\mathbf{S}$, $\mathbf{D}^{'}=\frac{1}{2}(\nabla \mathbf{u} + (\nabla \mathbf{u})^T) -\frac{1}{3}(\nabla.\mathbf{u}) \mathbf{I} $ is the deviatoric strain-rate tensor, $\mathbf{W}=\frac{1}{2}(\nabla \mathbf{u} - (\nabla \mathbf{u})^T)$ is the spin tensor, $\mathbf{D}_{pl}$ is the plastic component of the deviatoric strain-rate tensor, and G is the shear modulus of the material.
Eq. \ref{eq:S_evolution} for the evolution of deviatoric stress $\mathbf{S}$ is solved using a two-step operator-splitting algorithm \cite{sambasivan2013impact}. First, the deviatoric stress is evolved assuming a purely elastic deformation [i.e., setting the plastic component of the deviatoric strain-rate tensor, $\mathbf{D}_{pl}$ in Eq. \ref{eq:S_evolution}] in a predictor step: 
\begin{equation}
\frac{\partial \rho \mathbf{S}}{\partial t} +\nabla .(\rho \mathbf{S} \otimes \mathbf{u}) + \rho(\mathbf{S}\mathbf{W}-\mathbf{W}\mathbf{S}) -2\rho G \mathbf{D}^{'}=0
\label{eq:S_pred}
\end{equation}

This is followed by a correction step to remap the predicted stress onto the yield surface using the radial return algorithm [7-9]. In this step, $\mathbf{D}_{pl}$  is modeled assuming isotropic $J_2$ plasticity, with a von Mises yield criterion and associated flow-rule. The consistency condition is enforced explicitly and $\mathbf{D}_{pl}$ is computed using explicit plastic integration [ref] where the deviatoric stress is corrected to conform to the yield surface through the radial return algorithm \cite{ponthot2002unified, wilkins1963flow, huang2021explicit}.

\subsection{Pressure and Temperature equation of state}
The Mie-Gruneisen equation of state relates hydrostatic pressure to specific energy and density \cite{zeldovich2002shock} as shown below,
\begin{equation}
    p=p_c(\rho) + \Gamma \rho (e-e_c(\rho))
    \label{eq:EOS}
\end{equation}

in which $e_c$ is the cold compression energy, $p_c$ is the corresponding cold pressure, evaluated at 0 K. In Eq. \ref{eq:EOS} the “cold” (athermal) part of the specific internal energy $e_c$ is obtained from:

\begin{equation}
    e_c=e_{c,hydro}+e_{el}+e_{c,pl}
    \label{eq:ec}
\end{equation}
where the subscripts $hydro, el$, and $pl$ denote respectively hydrodynamic, elastic, and plastic contributions. The hydrodynamic contribution to the athermal part of the internal energy $e_c$ is given by:
\begin{equation}
    \dot{e}_{c,hydro}=\frac{p_c(\rho)}{\rho^2}\dot{\rho}
    \label{eq:echydro}
\end{equation}
in which $p_c(\rho)$ is the cold compression contribution to pressure, $e_{c,hydro}$ is obtained from Eq. \ref{eq:echydro} by integration,
\begin{equation}
    {e}_{c,hydro}=e_0 + \int_{\rho_0}^{\rho} \frac{p_c(\rho)}{\rho^2} d{\rho}
    \label{eq:echydro2}
\end{equation}

where $e_0$ is the reference internal energy at $T_0$, taken as 0 kcal/kg.
The plastic energy $e_{c,pl}$ contribution to cold specific internal energy is due to plastic work and is obtained by solving:
\begin{equation}
    \dot{e}_{c,pl}=(1-\beta) \mathbf{S}:\mathbf{D}_{pl}=(1-\beta)S_{vm}\dot{\varepsilon}_{pl}
    \label{eq:ecpl}
\end{equation}
In the above equation, $\beta$ is the Taylor-Quinney parameter $S_{vm}$ is the effective (von Mises) stress and $\dot{\varepsilon}_{pl}$ is the effective plastic strain rate. In the last term on the R.H.S. of Eq. \ref{eq:ecpl}, the rate of plastic work $\mathbf{S}:\mathbf{D}_{pl}$ is written as $S_{vm}\dot{\varepsilon}_{pl}$ for coaxial plasticity (J2 plasticity with Drucker’s postulate \cite{batra2006continuum}). 

The elastic energy contribution $e_{el}$ is obtained from the difference between the total deviatoric stress work and plastic work,  by solving:
\begin{equation}
    \dot{e}_{el}=\mathbf{S}:\mathbf{D}_{el}= \mathbf{S}:\mathbf{D}^{'}-S_{vm}\dot{\varepsilon}_{pl}
\end{equation}

Temperature is computed from the calorific equation of state as:

\begin{equation}
    T=T_0 + (e-e_c)/c_v,
\end{equation}
where $e$ is computed from \ref{eq:E}, $T_0$ is the reference temperature (298 K), $c_v$ is the isochoric specific heat and the cold energy $e_c$ is given by Eq. \ref{eq:ec}

\begin{table}[ht]
\centering
\caption{RDX Material Properties}
\label{tab:RDX}
\renewcommand{\arraystretch}{2}
\begin{adjustbox}{width=0.9\textwidth}
\begin{tabular}{|C{2.2cm}|C{6.2cm}|C{3.5cm}|C{0.8cm}|C{3.5cm}|}
\hline
Model & Expression & Parameters & Ref & Method  \\
\hline
EOS
(Pressure (Pa)) & $
 p_c(\rho) = \frac{3}{2} K_0 \left[ \left( \frac{\rho}{\rho_0} \right)^{7/3} - \left( \frac{\rho}{\rho_0} \right)^{5/3} \right]
\times \left[ 1 + \frac{3}{4} (K_0' - 4) \left\{ \left( \frac{\rho}{\rho _0} \right)^{2/3} - 1 \right\} \right] $ & $K_0 = 13 \times 10^9 \ \text{Pa}$ $\quad K_0' = 9.2$  
 & \cite{munday2011simulations}  & MD \\
\hline
Shear Modulus (Pa) & $G(P, T) = G_0 + a_1 P + a_2 (T - T_0)$ & \makecell[l]{
$G_0 = 5.314 \times 10^9$ Pa \\
$a_1 = 3.3774$ \\
$a_2 = -10.356 \times 10^6$ \\
\qquad \qquad Pa$\cdot$K$^{-1}$} & \cite{sen2024johnson} & CP informed by MD \cite{izvekov2021bottom}\\
\hline
Gruneisen coefficient & $\Gamma = \Gamma_0 + \gamma_1 \left( \frac{\rho_0}{\rho} \right) + \gamma_2 \left( \frac{\rho_0}{\rho} \right)^2
$  & \makecell[l]{
$\Gamma_0 = 0.667$ \\
$\gamma_1 = 2.00878$ \\
$\gamma_2 = -0.805669$
} & \cite{cawkwell2016eos} & MD \\
\hline
Johnson-Cook Strength Model & $S^y = \left( A + B \, \epsilon_{ps}^n \right)
\left[ 1 + C \ln \left( \frac{\dot{\epsilon}_{ps}}{\dot{\epsilon}_0} \right) \right] \times 
\left[ 1 - \left( \frac{T - T_{\text{ref}}}{T_{\text{melt}} - T_{\text{ref}}} \right)^m \right] $ & \makecell[l]{
$A = 0.3 \times 10^9$ Pa \\
$B = 0.1 \times 10^9$ Pa \\
$m = 3$ \\
$n = 0.1$ \\
$C = 1.8$ \\
$\dot{\epsilon}_0 = 4.36 \times 10^4$ s$^{-1}$
}
& \cite{sen2024johnson} & CP informed by experiments \cite{dick2004elastic}\\
\hline
Melt curve & $T_m(P) = T_{m,\text{ref}} \left[ 1 + \frac{p - p_{\text{ref}}}{a} \right]^{1/c}$ & \makecell[l]{
$T_{m,\text{ref}} = 478$ K \\
$P_{\text{ref}} = 0.0001$ GPa \\
$a = 0.9631$ GPa \\
$c = 2.8855$
} & \cite{kroonblawd2022rdx} & MD \\
\hline
Heat Capacity& $c_v=1980 \, J kg^{-1} K^{-1}$ & N/A &  \cite{sorescu2022heat}& MD \\
\hline
Thermal conductivity & \qquad $\chi=0.178 \quad Wm^{-1} K^{-1}$ & N/A & \cite{perriot2021pressure} & MD, averaged over $<100>$, $<010>$, and $<001>$ crystal orientations \\

\hline
\end{tabular}
\end{adjustbox}
\end{table}

\end{document}